\documentclass{article}
\usepackage[preprint]{colm2025_conference}
\setcitestyle{square,numbers}
\usepackage{multicol}

\usepackage{microtype}
\usepackage{makecell}
\usepackage{subcaption} 
\usepackage{longtable}
\usepackage[T1]{fontenc}    
\usepackage{tablefootnote}
\usepackage{threeparttable}
\usepackage{wrapfig}

\definecolor{green}{RGB}{0,150,10}
\definecolor{blue}{RGB}{0,148,181}
\definecolor{orange}{RGB}{194,153,107}
\usepackage[colorlinks,linkcolor=red,anchorcolor=green,citecolor=blue,urlcolor=MidnightBlue,]{hyperref}
\definecolor{HardBlue}{RGB}{0,45,120}
\hypersetup{
  colorlinks=true,
  urlcolor=HardBlue,
}
\usepackage{url}            
\usepackage{lmodern}
\usepackage{enumitem}

\usepackage{lineno}
\usepackage{tabularx}

\usepackage{booktabs}       
\usepackage{amsfonts}       
\usepackage{nicefrac}       
\usepackage{xcolor}         
\usepackage{times}
\usepackage{soul}
\usepackage{url}

\definecolor{background-grey}{RGB}{220,220,220}
\definecolor{cell-green}{RGB}{221, 255, 225}  
\definecolor{cell-red}{RGB}{255, 224, 224}  
\definecolor{light-green}{HTML}{A2D9A2}
\definecolor{llight-green}{HTML}{C7EFCF}
\definecolor{light-red}{HTML}{FFD1D1}
\definecolor{light-orange}{HTML}{FFC9A3}

\usepackage{float}
\usepackage{amsmath}
\usepackage{marvosym}
\usepackage{graphicx}
\usepackage{colortbl}
\usepackage{multirow}
\usepackage{subcaption}
\usepackage{changes}
\usepackage{bookmark}
\usepackage{listings}
\lstdefinelanguage{Dialogue}{
  morekeywords={Influencer,Voter,rating},
  sensitive=false,
  morecomment=[l]{//},
}

\usepackage{bibentry}
\nobibliography*

\usepackage{mdframed}
\usepackage{caption}
\usepackage{fancyhdr}
\usepackage{datetime}
\usepackage{adjustbox}
\usepackage{amssymb}
\usepackage{tcolorbox}
\usepackage{makecell}
\usepackage{multirow}      
\usepackage{array}         
\usepackage{fancyvrb}
\usepackage{fvextra}
\usepackage{pifont}
\usepackage{tikz}
\usepackage{wasysym}
\usepackage{xspace}
\usepackage{setspace}
\usepackage{algorithm, algorithmic}
\usepackage{soul} 

\usepackage{hhline}

\definecolor{evidbgcolor}{HTML}{FFE6E6}
\definecolor{stepbgcolor}{HTML}{F0F0F0}
\definecolor{evidfgcolor}{HTML}{CC0000}

\usepackage{ulem}     

\definecolor{lightred}{RGB}{255,200,200}

\tcbuselibrary{listings,breakable}

\usepackage{geometry}
\definecolor{Blue4Head}{RGB}{58,104,153}

\title{%
\begin{center}
  \begin{minipage}[c]{0.19\textwidth}
    \flushright
    \includegraphics[width=1.2cm]{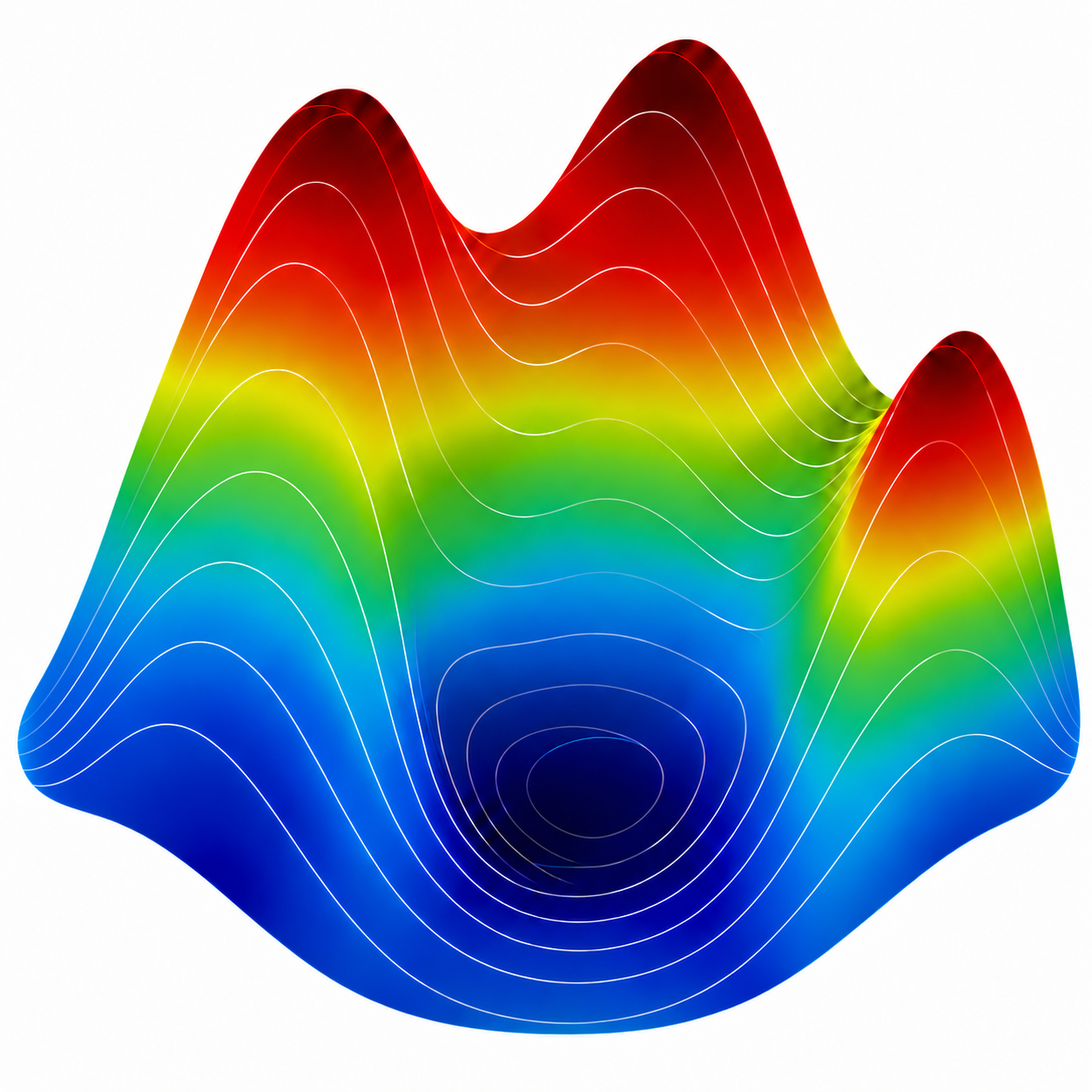} \quad
  \end{minipage}
  \begin{minipage}[c]{0.75\textwidth}
    \raggedright
    \Large\bfseries
    \opensourceabbf{}: \methodname{} \\ Are Scalable Architectures
  \end{minipage}
\end{center}
}

\def\methodname{Energy-based Foundation Models}
\def\methodabbf{EBFM}
\def\opensourceabbf{OpenEBFM}
\definecolor{lightred}{RGB}{247, 220, 220}

\author{%
  \makebox[\textwidth][c]{%
    \begin{tabular}{@{}c@{}}
      Yuan Pu$^*$\textsuperscript{\S},
      Yudong Lu$^*$,
      Ting Wang$^*$,
      Zhicong Hong\textsuperscript{\S},
      Qinuo Li,
      \tabularnewline
      Xuansheng Wu,
      Na Zou,
      Xia Hu,
      Guanchu Wang
    \tabularnewline[5pt]
    \normalfont{Shanghai Artificial Intelligence Laboratory}
    \tabularnewline[5pt]
    \normalfont\texttt{\{puyuan,luyudong,wangting1,hongzhicong,liqinuo,}
    \tabularnewline
    \normalfont\texttt{wuxuansheng,zouna,huxia,wangguanchu\}\@@pjlab.org.cn}
    \end{tabular}%
  }%
}

\renewcommand{\thefootnote}{\fnsymbol{footnote}}
\begin{document}
    
\hypersetup{
    linkcolor=black,
    filecolor=black,      
    urlcolor=blue,
    citecolor=blue,
}

\maketitle

\begingroup
\renewcommand{\thefootnote}{\fnsymbol{footnote}}
\footnotetext[1]{Equal contribution to this work.}
\footnotetext[4]{Work completed while at Shanghai Artificial Intelligence Laboratory.}
\endgroup

\begin{center}
    \vspace{-8mm}
    \includegraphics[height=1.5em]{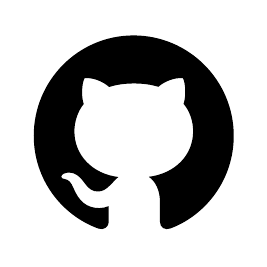}
    \href{https://github.com/datamllab/openebfm}{\textsf{\bfseries{\textcolor{HardBlue}{https://github.com/datamllab/openebfm}}}}
    \\
    \includegraphics[height=1.5em]{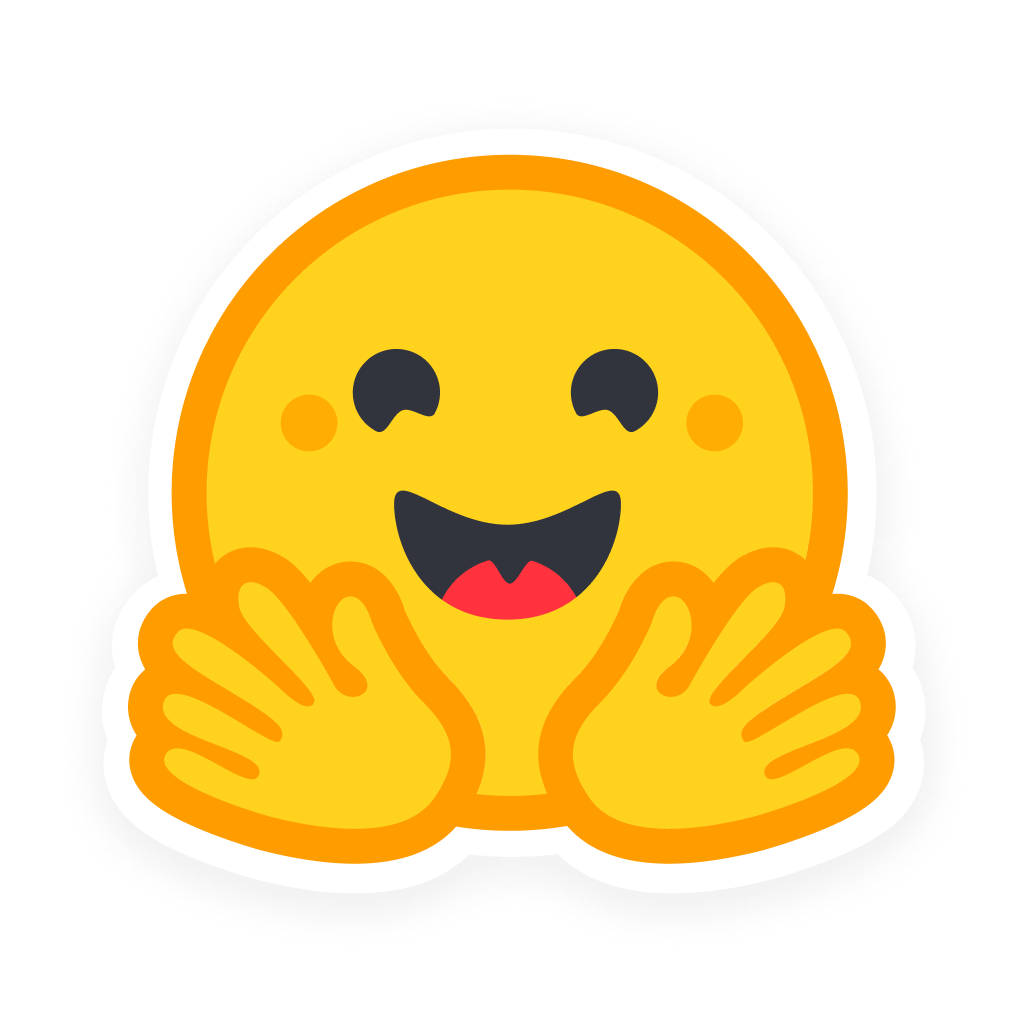}
    \href{https://huggingface.co/collections/guan-wang/openebfm}{\textsf{\bfseries{\textcolor{HardBlue}{https://huggingface.co/collections/guan-wang/openebfm}}}}
    \vspace{2mm}
\end{center}



\begin{abstract}
\begin{center}
\begin{minipage}{0.7\textwidth}
Energy-based models have shown promising properties for domain-specific tasks but remain challenging in modeling general-purpose scenarios.
To bridge this gap, we design Energy-Based Foundation Models~(\methodabbf{}s) for general language modeling.
\methodabbf{}s integrate energy steering and token generation within a single architecture to leverage the energy-guided search to refine token generation.
Specifically, \methodabbf{}s first optimize a latent state over the energy landscape, then leverages the refined latent state for token generation.
By performing such energy-guided pre-training on large-scale corpora, \methodabbf{}s learn generally effective energy landscape where favorable tokens correspond to lower energy values.
To explore the scaling behavior of \methodabbf{}s, we pre-train \methodabbf{}s on three large-scale corpora, then evaluate across 26 downstream tasks spanning a broad range of domains.
Our results demonstrate that \methodabbf{}s have promising scaling behaviors with increasing model size and training data, and outperform GPTs on most downstream tasks.
These findings highlight \methodabbf{}s as a promising architecture for large-scale pre-training and general purpose problem solving.
We fully release the model weights and training infrastructure to enable the community to completely reproduce and extend our results.
\end{minipage}
\end{center}
\end{abstract}

\section{Introduction}

Energy-based models (EBMs) show potential in modeling complex dependencies and constraints in problem-solving~\cite{lecun2006tutorial,huembeli2022physics,du2019implicit}.
Instead of directly parameterizing a normalized probability distribution, EBMs assign energy values to each possible solution, and learn an energy landscape such that desirable solutions receive low energy and undesirable ones receive high energy~\cite{carbone2024hitchhiker,song2021train,mordatch2018concept,du2022learning}.
This formulation has shown promise in several domains, including pattern recognition~\cite{xie2019learning,zhijian2024energy}, structured prediction~\cite{gladstone2024cognitively}, robot controlling~\cite{boney2020regularizing,benzi2022energy}, and sequential planning~\cite{gkanatsios2023energy}.
These applications indicate that energy-based modeling offers an effective alternative paradigm for learning and inference~\cite{duvenaud2020your}.

%


However, existing EBMs are typically designed and evaluated within a single task or domain, such as representation learning over specific data distributions~\cite{du2022learning}, structured understanding and prediction for particular tasks~\cite{liu2024vision,qi2023energy}, or control and planning in specialized environments~\cite{benzi2022energy,boney2020regularizing}.  
As a result, EBMs remain specialized to particular domains, while their generalizability across diverse tasks and domains remains largely unexplored.
In contrast, advanced foundation models, including generative language models~(GPT)~\cite{radford2019language} and diffusion language models~(DLM)~\cite{sahoo2024simple}, have demonstrated strong generalizability across heterogeneous tasks and contexts~\cite{radford2019language,touvron2023llama,sahoo2024simple,nie2026large}. 
Whether EBMs can achieve such general-purpose capabilities has yet to be established.
Therefore, it remains an open and fundamental question: \textbf{\textit{Can EBMs serve as foundation models for general-purpose tasks?}}


Scaling EBMs to general-purpose tasks is non-trivial problem. 
A fundamental challenge is that traditional EBMs are designed to estimate scalar energy values rather than directly generating structured outputs~\cite{lecun2006tutorial, song2021train, bond2021deep}. 
Therefore, it remains challenging to leverage the energy evaluation capability of EBMs to improve token generation in foundation models~\cite{du2019implicit, gladstone2025energy}. 
To address this challenge, we design the \methodname{}~(\methodabbf{}s), a novel architecture that integrates energy steering with token generation to enable general-purpose capability in EBMs. 
Specifically, \methodabbf{}s have two complementary components: an energy head and a generative head. 
During inference, the energy head estimates the energy of the input context and iteratively steers a latent variable toward lower-energy regions. 
After this energy-steering step, the generative head takes the steered variable and context embedding to predict new tokens.
In this way, \methodabbf{}s significantly bridge the gap between energy evaluation and generative modeling, enabling the model to acquire broad capabilities through large-scale pre-training.
Moreover, by refining the latent representations toward lower-energy states, \methodabbf{}s can better capture contextual dependencies and consistency in large-scale corpora, improving their generalization to diverse general-purpose tasks.

\begin{wrapfigure}{r}{0.5\linewidth}
\vspace{-4mm}
\centering
\begin{subfigure}[t]{0.24\textwidth}
        \centering
        \includegraphics[width=\linewidth]{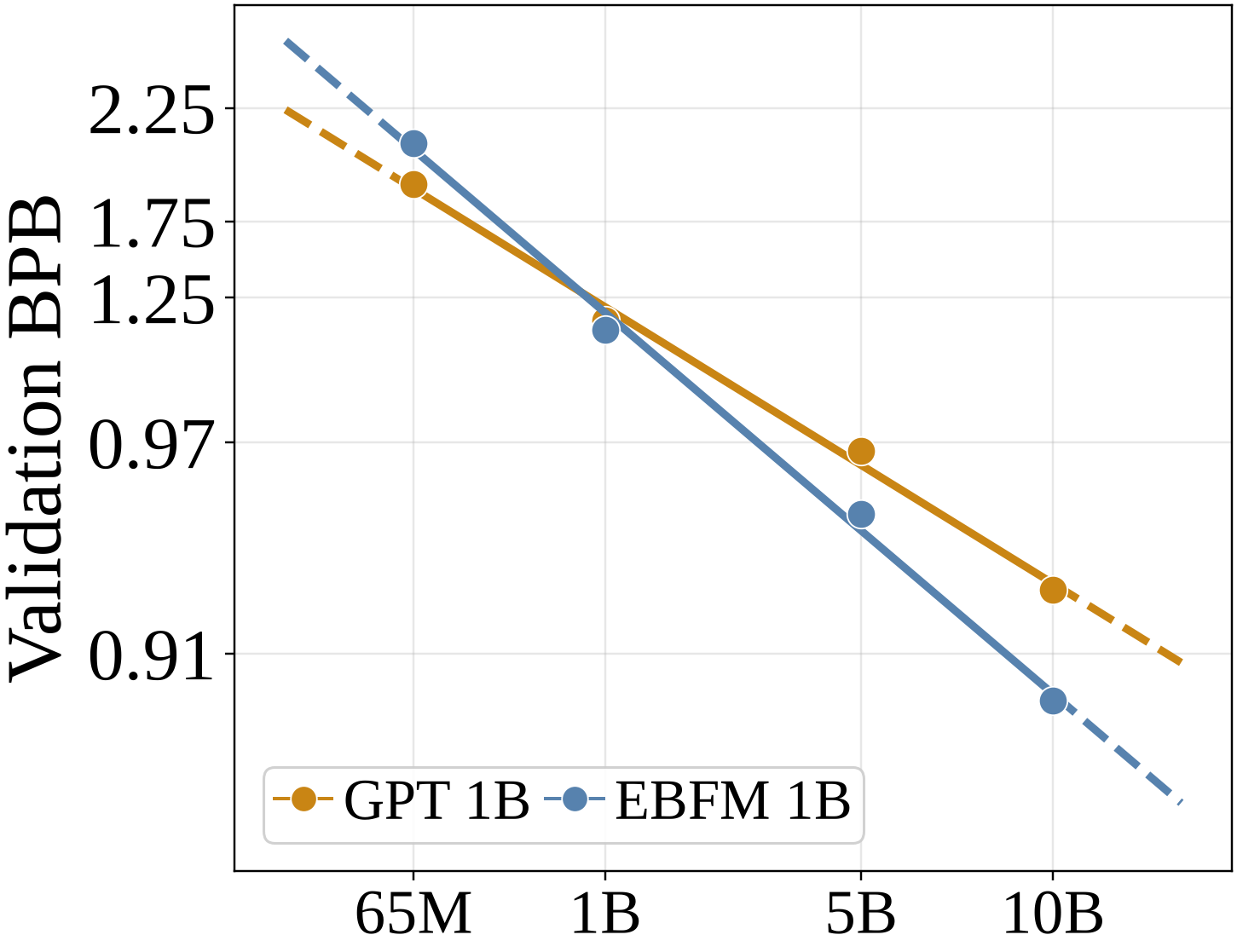}
        \caption{Training tokens} 
    \end{subfigure}
\begin{subfigure}[t]{0.24\textwidth}
    \centering
    \includegraphics[width=\textwidth]{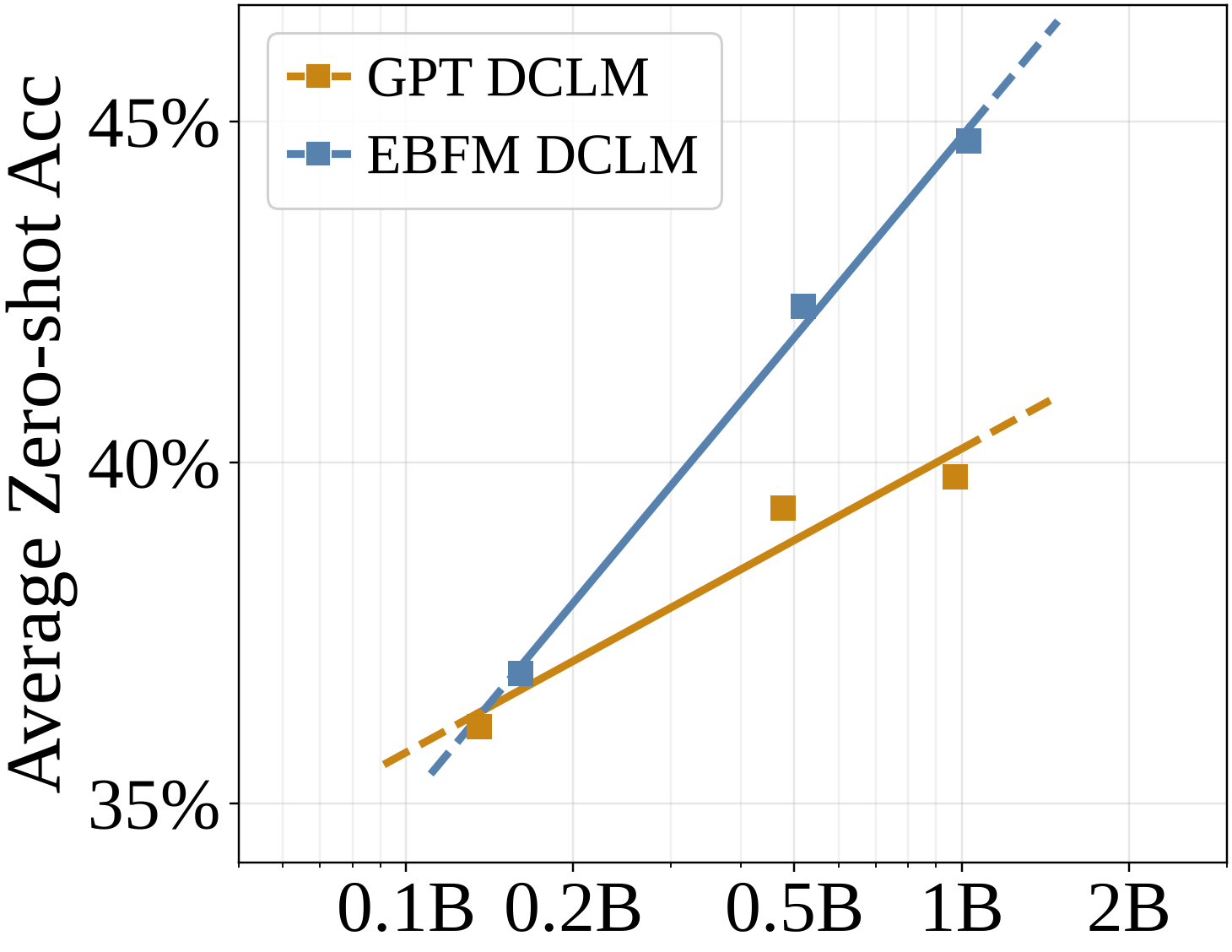}
    \caption{Model size}
\end{subfigure}
\caption{\label{fig:advantage}\footnotesize Scaling behavior of \methodabbf{}s on the DCLM dataset.}
\end{wrapfigure}


Beyond designing the architecture of \methodabbf{}, we systematically evaluate its scaling relation between the numbers of its parameters and training samples and its downstream task performs to demonstrate its efficiency on using training data against other architectures.
Specifically, we pre-train \methodabbf{}s on three large-scale corpora~\cite{penedo2024fineweb, diao2026nemotron, li2024datacomp}, and the results demonstrate promising scaling behavior with increasing model size and training data, as shown in Figure~\ref{fig:advantage}.
In addition, we evaluate \methodabbf{}s across 26 downstream tasks spanning a broad range of domains under zero-shot settings, and exhibit consistent advantages over GPTs.
These findings establish \methodabbf{}s as a promising architecture for large-scale pre-training and general-purpose problem solving.  
Finally, we fully release the training recipe, infrastructure, and other details, enabling the community to completely reproduce and extend our results.

To summarize, the contributions of this work are as follows:
\begin{itemize}

    \item \textbf{\methodname{}.} We design \methodabbf{}s, an energy-based  architecture for foundation language modeling.
    \methodabbf{}s leverage energy-steering to refine token generation, capturing contextual dependencies more effectively.    
    
    
    
    
    \item \textbf{Scaling Law.} \methodabbf{}s exhibit favorable scaling behavior with increasing model size and training data, compared with Generative Pre-trained Transformers~(GPTs), and Masked Diffusion Language Models~(MDLMs).

    \item \textbf{General-Purpose Task Solving.} \methodabbf{}s exhibit promising zero-shot performance across 26 downstream tasks spanning a broad range of domains, as measured by validation error and accuracy.
    
    \item \textbf{Open-source Ecosystem.} We fully release the model weights~(\opensourceabbf{}), training recipe, and infrastructure, such that the community can completely reproduce and extend this work.
    
\end{itemize}









\section{Preliminary}

\subsection{Notations}

\begin{figure}[t]
\centering

\begin{subfigure}{0.5\linewidth}
    \centering
    \includegraphics[width=\linewidth]{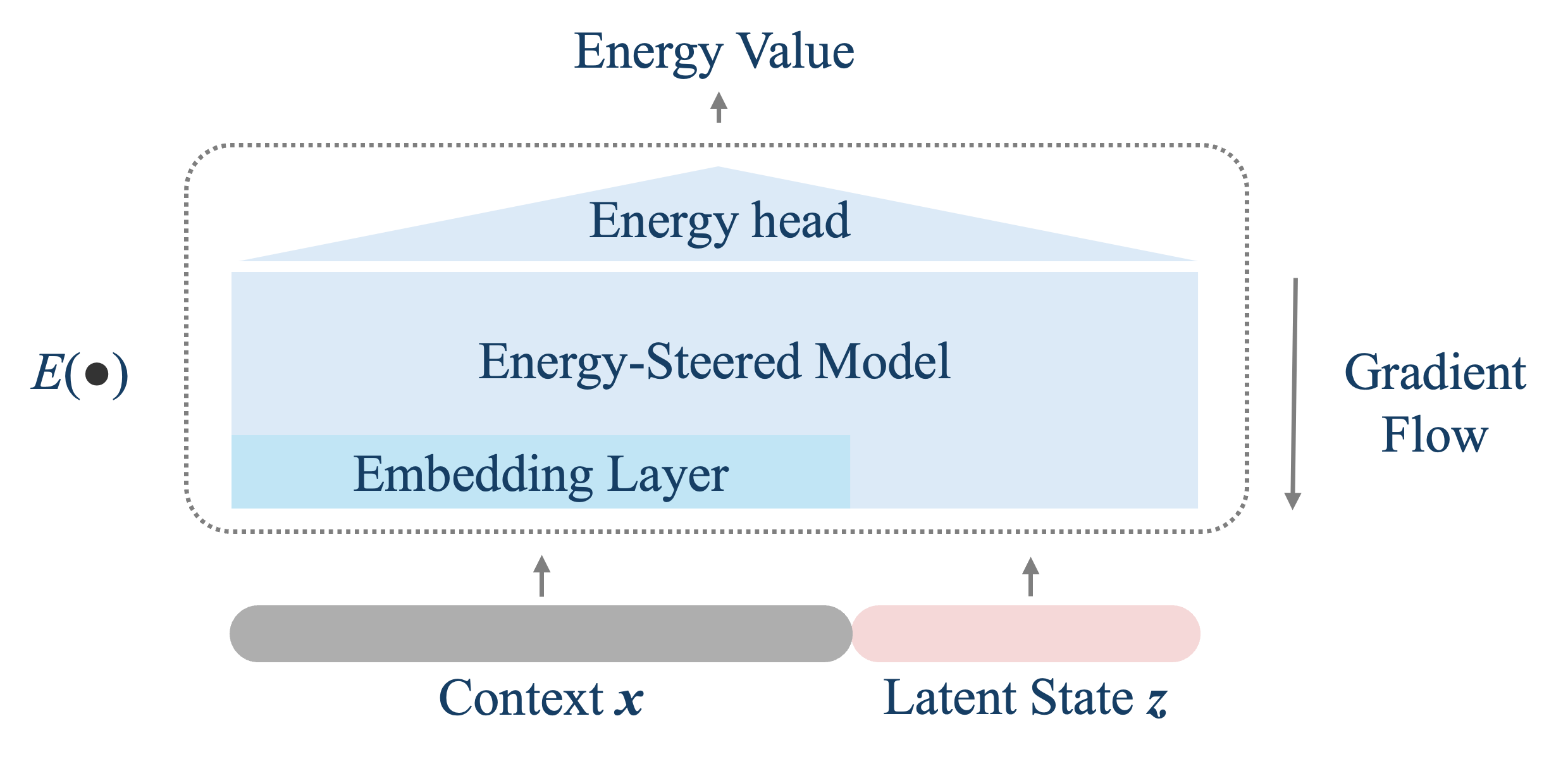}
    \caption{Energy steering}
\end{subfigure}
\quad \quad
\begin{subfigure}{0.4\linewidth}
    \centering
    \includegraphics[width=\linewidth]{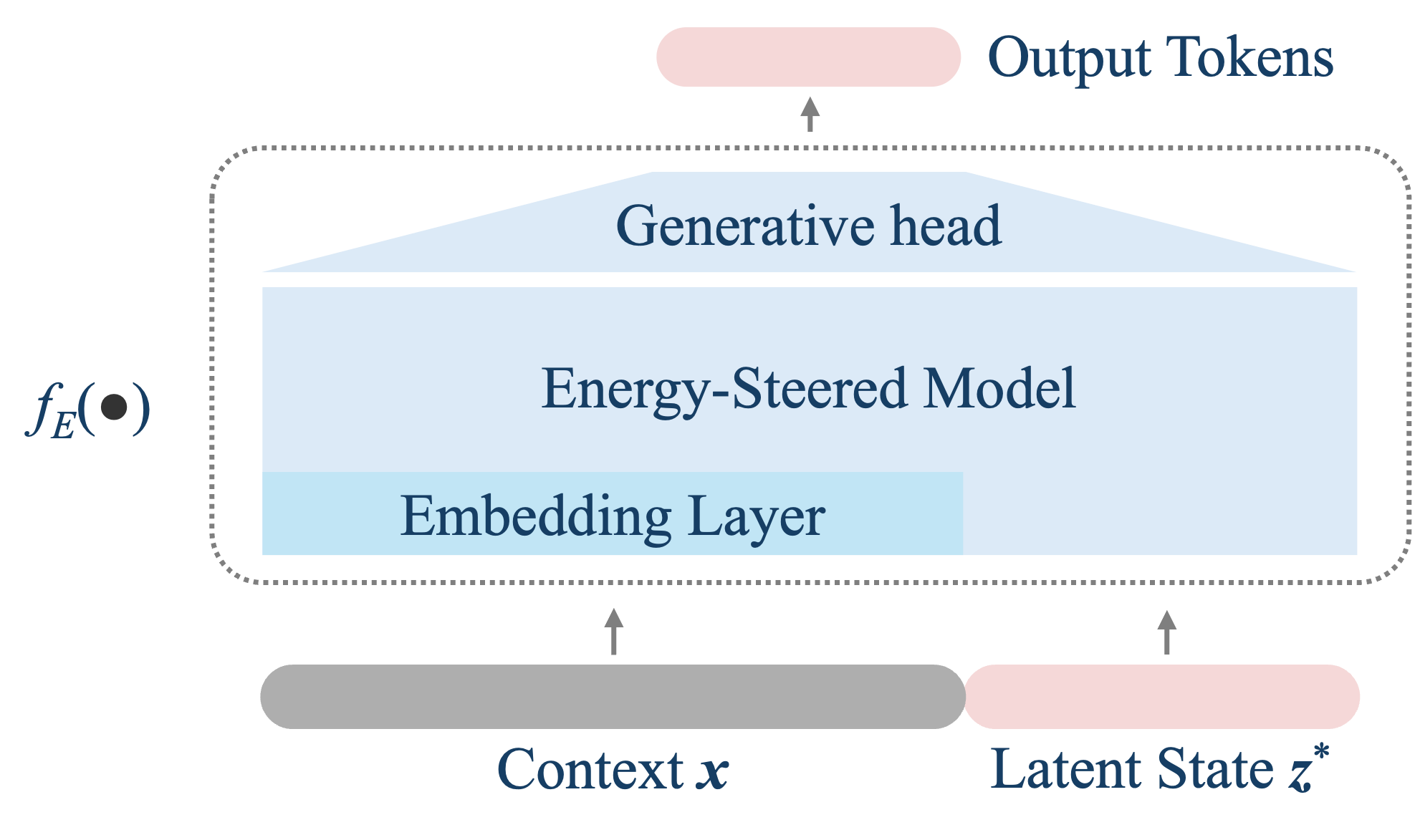}
    \caption{Token generation}
\end{subfigure}

\caption{
Generation process of \methodabbf{}s.
(a) \textbf{Energy steering:} The energy head estimates the energy of the latent state and backpropagates the energy gradient with respect to $\boldsymbol{z}$ to obtain a low-energy latent state $\boldsymbol{z}^*$ for token generation.
(b) \textbf{Token generation:} The generative head uses the optimized latent state $\boldsymbol{z}^*$ to generate the next token.
}
\label{fig:ebm_generation}
\end{figure}

We specify the notation in this work.
The overall framework of \methodabbf{} is given in Figure~\ref{fig:ebm_generation}.
Specifically, \methodabbf{} takes two inputs: a context sequence $\boldsymbol{x} \in \mathcal{X}$ and a latent variable $\boldsymbol{z} \in \mathcal{R}^{D}$, where $\mathcal{X}$ denotes the sequence space and $D$ denotes the embedding dimension.
Given these inputs, \methodabbf{} performs both energy evaluation and token generation based on an energy head and a generative head, respectively.
As in Figure~\ref{fig:ebm_generation}, the energy head is a linear layer for energy evaluation, while the generative head is an embedding layer with \texttt{softmax} operations for token generation.
Both heads share a unified Transformer backbone, which enables energy evaluation and token generation within a single architecture.
Let $E(\boldsymbol{x}, \boldsymbol{z} ~|~ \theta) \in \mathbb{R}$ denote the forward pass through the energy head, which outputs an energy score, as shown in Figure~\ref{fig:ebm_generation}~(a).
Let $f_E(\boldsymbol{x}, \boldsymbol{z} ~|~ \theta) \in \mathbb{V}$ denote the forward pass through the generative head, which outputs tokens from the vocabulary space $\mathbb{V}$, as illustrated in Figure~\ref{fig:ebm_generation}~(b).
We develop pre-training processes such that energy evaluation helps refine token generation across general-purpose scenarios.




\subsection{Energy-based Models}

Energy-based models~(EBMs) focus on learning an energy landscape over the solution space, assigning low energy to desirable solutions and high energy to undesirable ones.
However, applying \methodabbf{}s as foundation models to general-purpose tasks remains challenging, because they naturally serve as critics that evaluate candidate solutions rather than directly generating outputs~\cite{lecun2006tutorial, song2021train, bond2021deep}. 
Bridging this gap is non-trivial, as fully enumerating the token space to identify the lowest-energy solution is computationally infeasible.
Our proposed \methodabbf{}s address this challenge by leveraging energy evaluation to refine token generation, thus enabling general-purpose task solving.









\section{\methodname{}}

We design the architecture of \methodabbf{}s in this section, including \methodabbf{} inference and training approaches.

\subsection{\methodabbf{} Inference}

As shown in Figure~\ref{fig:ebm_generation}, \methodabbf{} inference consists of two key steps: \textbf{\textit{energy steering}} and \textbf{\textit{token generation}}.
Energy steering captures the contextual dependencies to refine token generation.

\paragraph{Energy Steering.}
\methodabbf{}s perform energy steering to refine a latent state $\boldsymbol{z}$ to maximize contextual consistency.
Intuitively, $\boldsymbol{z}$ serves as a controllable representation that influences the subsequent tokens.
Given $\boldsymbol{x}$ and $\boldsymbol{z}$, the energy head $E(\boldsymbol{x}, \boldsymbol{z}\mid\theta)$ evaluates the compatibility between $\boldsymbol{x}$ and $\boldsymbol{z}$, where lower energy indicates a better latent state under the current context. 
Energy steering aims to identify a low-energy latent state $\boldsymbol{z}^*$ that effectively captures the contextual dependencies required for subsequent token generation.
Formally, the energy steering objective is given as follows:
\begin{align}
\label{eq:energy_steer}
\boldsymbol{z}^* = \arg\min_{\boldsymbol{z}\in \mathcal{R}^D} E(\boldsymbol{x}, \boldsymbol{z} ~|~ \theta)
\end{align}

Note that the exact solution of Equation~(\ref{eq:energy_steer}) is intractable, \methodabbf{} approximates it through a $K$-step Markov chain Monte Carlo~(MCMC) sampling.
Specifically, MCMC sampling starts from an initial latent variable sampled from a standard Gaussian distribution $\boldsymbol{z}^{0} \sim \mathcal{N}(\boldsymbol{0},\mathbf{I})$, then update the latent variable through gradient descent as follows:
\begin{align}
\label{eq:mcmc}
\boldsymbol{z}^{k} = \boldsymbol{z}^{k-1} - \alpha \nabla_{\!\!\boldsymbol{z}^{k-1}} E(\boldsymbol{x}, \boldsymbol{z}^{k-1} ~|~ \theta),
\end{align}
where $\alpha$ controls the step size.
We omit the stochastic noise commonly used in the MCMC sampling, as we empirically observe that deterministic updates yield more stable optimization in large-scale pre-training.
During energy steering, the model parameters $\theta$ remain fixed, and gradients are computed only with respect to the latent state $\boldsymbol{z}$.
After $K$ step refinement, the optimized latent state is approximated by the final state, i.e.,
$\boldsymbol{z}^{*}\approx\boldsymbol{z}^{K}$.





\paragraph{Token Generation.}
\methodabbf{} leverages the low-energy latent state $\boldsymbol{z}^*$ to generate subsequent tokens.
Specifically, given the context $\boldsymbol{x}$, the generative head $f_E(\bullet~|~ \theta)$ produce new tokens as follows:
\begin{align}
\hat{\boldsymbol{y}} \sim f_E(\boldsymbol{x}, \boldsymbol{z}^*~|~ \theta),
\end{align}
where $f_E(\boldsymbol{x}, \boldsymbol{z}^*~|~ \theta)$ produces the distribution of new tokens over the entire vocabulary space.
We only consider text data in this work, where $\boldsymbol{x}$ and $\boldsymbol{y}$ are text tokens.


\begin{figure}
    \centering
    \includegraphics[width=0.9\linewidth]{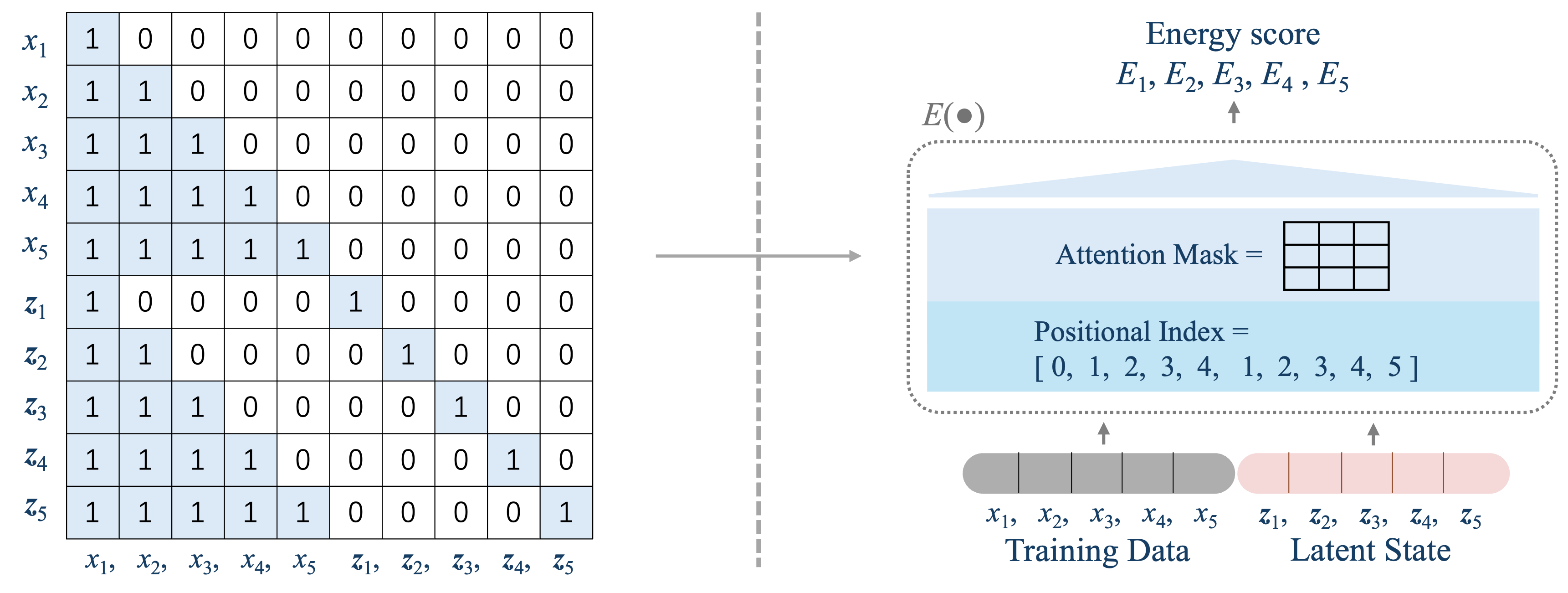}
    \caption{\methodabbf{}s parallelize energy steering for training tokens at all positions through re-designed attention masks and positional indexes. \textbf{Left Panel:} the re-designed attention mask where 0 indicates blocking the cross-token-attention. \textbf{Right Panel:} the forward path with re-designed attention mask and positional index.}
    \label{fig:attention-mask}
\end{figure}

\subsection{\methodabbf{} Training}


\methodabbf{}s parallelize energy steering and token generation during training to improve training efficiency.


\paragraph{Energy Steering.}
\methodabbf{}s parallelize the energy steering at the sequence level for efficient large-scale pre-training.
Specifically, for an input sequence $\boldsymbol{x}_{[1:S]} = [x_1,\ldots,x_S]\in\mathcal{X}$, \methodabbf{}s follow the causal attention paradigm such that latent state $\boldsymbol{z}_i$ can only attend to its prefix tokens $\boldsymbol{x}_{[1:i]}$.
Therefore, the steering process for $\boldsymbol{z}_i$ is given by 
\begin{equation}
    \boldsymbol{z}^*_i = \arg\min_{\boldsymbol{z}\in \mathcal{R}^D} E(\boldsymbol{x}_{[1:i]}, \boldsymbol{z} ~|~ \theta).
\end{equation}

However, separately steering each latent state $\boldsymbol{z}_1, \cdots, \boldsymbol{z}_S$ results in significant computational overhead during large-scale pre-training. 
To improve efficiency, \methodabbf{}s perform the steering for the latent states at all positions in parallel.
To be concrete, \methodabbf{}s concatenate the training data $\boldsymbol{x}_{[1:S]}$ with latent states $\boldsymbol{z}_{[1:S]}$ into $[\boldsymbol{x}_{1:S}, \boldsymbol{z}_{1:S}]$, and simultaneously estimates energy values $E_1, \cdots, E_S$ corresponding to $[\boldsymbol{x}_{[1:1]}, \boldsymbol{z}_1], \cdots [\boldsymbol{x}_{[1:S]}, \boldsymbol{z}_S]$ within a single forward pass. 
The left panel of Figure~\ref{fig:attention-mask} illustrates the attention mask for the concatenated sequence $[\boldsymbol{x}_{[1:5]}, \boldsymbol{z}_{[1:5]}]$, which restricts each latent state $\boldsymbol{z}_i$ to attend only to its prefix tokens $\boldsymbol{x}_{[1:i-1]}$ and itself.
This design preserves the causal dependency required for language modeling while allowing all latent variables to be processed in parallel during training.

Beyond the attention masks, the positional indices of $\boldsymbol{z}_1, \boldsymbol{z}_2, \cdots, \boldsymbol{z}_S$ are assigned as $[1, 2, 3, \cdots, S]$ to align each latent state with its corresponding prediction position.
These designs enable \methodabbf{}s to efficiently perform energy steering for all sequence positions in parallel, which follows a sequence-level optimization objective given by
\begin{equation}
    \boldsymbol{z}^*_{[1:S]} = \arg\min_{\boldsymbol{z} \in \mathcal{R}^{S \times D}} \mathrm{sum}_{1\leq i \leq S} \!\left( E_i(\boldsymbol{x}_{[1:S]}, \boldsymbol{z}_{[1:S]} ~|~ \theta) \right).
\end{equation}
\methodabbf{}s thereby approximate the solution through $K$-step MCMC sampling:
\begin{equation}
\label{eq:MCMC-seq}
    \boldsymbol{z}^k_{[1:S]} = \boldsymbol{z}^{k-1}_{[1:S]} - \alpha \nabla_{\!\!\boldsymbol{z}^{k-1}_{[1:S]}} \mathrm{sum}_{1\leq i \leq S} \!\left( E_i(\boldsymbol{x}_{[1:S]}, \boldsymbol{z}_{[1:S]} ~|~ \theta) \right),
\end{equation}
where $\boldsymbol{z}^0_{[1:S]} \sim \mathcal{N}(\boldsymbol{0}, \mathbf{I})$.
The final state serves as an approximation to the optimal solution $\boldsymbol{z}^{*}_{[1:S]}\approx\boldsymbol{z}^{K}_{[1:S]}$.







\paragraph{Supervised Training.}
Given the input sequence $\boldsymbol{x}_{[1:S]}$ from the training corpora and optimized latent state $\boldsymbol{z}^{*}_{[1:S]}$ from the energy steering, \methodabbf{}s predict the subsequent token probabilities through the generative head $p(\boldsymbol{x}_{[2:S+1]}) = f_E(\boldsymbol{x}_{[1:S]}, \boldsymbol{z}^*_{[1:S]}; \boldsymbol{x}_{[2:S+1]} ~|~ \theta)$, and then update the model parameters $\theta$ to minimize the loss function:
\begin{align}
\label{eq:loss-function}
\mathcal{L}(\theta) = - \log f_E(\boldsymbol{x}_{[1:S]}, \boldsymbol{z}^*_{[1:S]}; \boldsymbol{x}_{[2:S+1]} ~|~ \theta).
\end{align}



\subsection{Algorithms}

\methodabbf{} inference and training are given in Algorithms~\ref{alg:inference} and~\ref{alg:training}, respectively.
Specifically, during inference, \methodabbf{}s perform energy steering~(line~3) on the latent state, which in turn refines token generation~(line~5).
During training, with each mini-batch of inputs, \methodabbf{}s perform energy steering~(line~2), estimate loss value~(line~3), and update model weights~(line 4).
Across both inference and training, energy steering is the key step that enables \methodabbf{}s to significantly capture in-context dependency for generating new tokens. 

\begin{minipage}{0.48\textwidth}
\begin{algorithm}[H]
\caption{\methodabbf{} Inference}
\label{alg:inference}
\textbf{Input:} input sequence $\boldsymbol{x}_{[1:S]}$. \\
\textbf{Output:} predicted next token $\hat{x}_{S+1}$.\\
\vspace{-4mm}
\begin{algorithmic}[1]


\STATE $\boldsymbol{z}^0 \sim \mathcal{N}(\boldsymbol{0}, \mathbf{I})$
\FOR{\emph{layer} $k := 1 \text{ to } K$}

\STATE Update latent state by Eq.~(\ref{eq:mcmc}) \textcolor{gray}{\# Energy steering}
\ENDFOR

\STATE $\hat{x}_{S+1} \sim f_E(\boldsymbol{x}, \boldsymbol{z}^*~|~ \theta)$ \textcolor{gray}{\# Token generation}

\RETURN $\hat{x}_{S+1}$
\end{algorithmic}
\end{algorithm}
\end{minipage}
\begin{minipage}{0.48\textwidth}
\begin{algorithm}[H]
\caption{\methodabbf{} Pre-training}
\label{alg:training}
\textbf{Input:} pre-training dataset $\mathcal{D}$. \\
\textbf{Output:} \methodabbf{} model weights $\theta^*$.\\
\vspace{-3mm}
\begin{algorithmic}[1]


\FOR{mini-batches $\boldsymbol{x} \sim \mathcal{D}$}

\STATE Update latent state by Eq.~(\ref{eq:MCMC-seq}) 
\STATE Estimate loss function $\mathcal{L}(\theta)$ given by Eq.~(\ref{eq:loss-function})
\STATE Update $\theta$ by $\theta^* = \arg \min \mathcal{L}(\theta)$
\ENDFOR

\RETURN $\theta^*$
\end{algorithmic}
\end{algorithm}
\end{minipage}

\section{Evaluation Results}

We pre-train \methodabbf{}s on large-scale corpora with billions of tokens. 
Evaluations of \methodabbf{}s comprehensively cover pre-training, supervised finetuning, zero-shot question-answering, zero-shot generations to assess their capabilities across diverse learning and reasoning tasks.
We compare \methodabbf{}s with established foundation architectures, including GPTs and Diffusion Language Models~(DLMs)~\cite{sahoo2024simple}.

\subsection{Pre-training and Evaluation Recipe}
\label{sec:recipe}


\paragraph{Pre-training Recipe.}
We pre-train \methodabbf{}s on three large-scale corpora, including Fineweb~\cite{penedo2024fineweb}, Climbmix~\cite{diao2026nemotron}, and DCLM~\cite{li2024datacomp}.
To investigate scaling with model size, we vary the number of layers as $d \in \{12, 20, 26\}$ and jointly scale the model depth and hidden width following established Transformer scaling practices.
Specifically, we follow the configuration of GPT-2 to have the hidden width $d_{\mathrm{model}}=\gamma d$, where $\gamma=64$, and number of attention heads $n_{\mathrm{head}}=d_{\mathrm{model}} / 128$.
Other hyper-parameter settings are provided in Table~\ref{eq:train-recipe}.
After the pre-training, \methodabbf{}s are evaluated with three strategies as follows.


\begin{table}[b]
    \centering
    \small
    \resizebox{\textwidth}{!}{
    \begin{tabular}{lc|lc|lc}
        \toprule
        Hyper-parameter & \methodabbf{} & Hyper-parameter & \methodabbf{} & Hyper-parameter & \methodabbf{} \\
        \midrule
        Tokenizer & Nanochat & MCMC Steps & 1 & Weight Decay & 0.2 \\
        Vocabulary Size & 32{,}768 & MCMC Step Size & 300 & Warmup Ratio & 0.0057 \\
        Context Length & 2{,}048 & $\alpha$ Learning Rate & 1.8 & Warmdown Ratio & 0.65 \\
        Global Batch Size & 512 & Muon Learning Rate & 0.02 & Muon Momentum & 0.95 \\
        Optimizer & Muon + AdamW & Embedding Learning Rate & 0.3 & Muon $\beta_2$ & 0.95 \\
        Gradient Clipping & 1& Final LR Fraction & 0.05 & AdamW $\beta_1$, $\beta_2$ & 0.8, 0.95 \\
        \bottomrule
    \end{tabular}
    }
    \caption{\label{eq:train-recipe}
    Hyper-parameter setting for \methodabbf{} pre-training.}
\end{table}

\paragraph{Scaling Law Evaluation.} 
Scaling behaviors are evaluated on the validation sets of Fineweb, Climbmix, and DCLM datasets.
Validation error is measured by bits per byte~(BPB) defined by $\text{BPB} = (\log_2 \text{PPL}) / (\text{Bytes per token})$, where PPL denotes the token-level perplexity on the validation datasets.
BPB normalizes the PPL by the number of bytes, enabling more meaningful comparisons across different tokenizers.
Lower BPB indicates better predictive performance.
We evaluate the scaling behavior of models by observing how rapidly their validation error decreases with increasing training data and model size.


\paragraph{Zero-shot QA.} 
Each foundation model is evaluated on 19 downstream tasks, including AGI-Eval LSAT-AR, ARC-Challenge, ARC-Easy, BIG-bench CS Algorithms, BIG-bench Dyck Languages, BIG-bench Language Identification, BIG-bench Operators, BIG-bench QA-Wikidata, BoolQ, CommonsenseQA, COPA, CoQA, HellaSwag, HellaSwag-ZeroShot, LAMBADA-OpenAI, OpenBookQA, PIQA, Winograd, and WinoGrande datasets. 
These tasks span diverse domains and capabilities, including science, algorithmic reasoning, language understanding, factual knowledge, logical reasoning, symbolic reasoning, and commonsense reasoning. 
Such broad evaluation provides a comprehensive assessment of the general-purpose capabilities of pretrained models.

\paragraph{Zero-shot Generation.} 
Each foundation model is pre-trained on the OpenWebText dataset~\cite{Gokaslan2019OpenWeb} and evaluated on seven downstream datsests, including PTB~\cite{marcus1993building}, Wikitext~\cite{merity2016pointer}, LM1B~\cite{chelba2013one}, Lambada~\cite{paperno2016lambada}, AG News~\cite{zhang2015character}, Pubmed, and Arxiv~\cite{cohan2018discourse} datasets.
Generation quality is measured by the validation BPB on each downstream dataset.

\begin{figure}[t]
    \centering
    \begin{subfigure}[t]{0.499\linewidth}
        \centering
        \includegraphics[width=\linewidth]{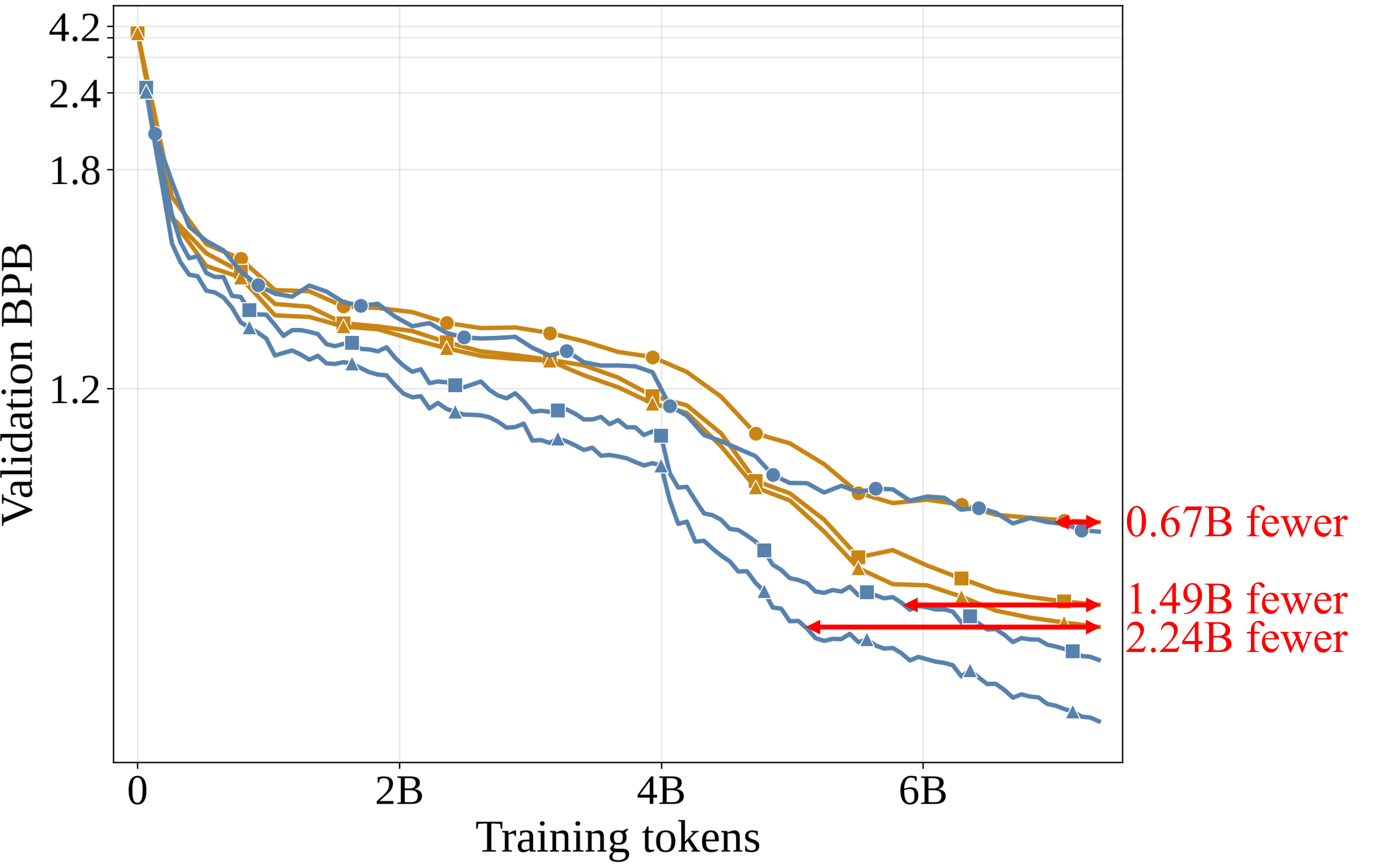}
        \caption{Climbmix}
        \label{fig:pretrain_val_bpb_climbmix}
    \end{subfigure}
    \begin{subfigure}[t]{0.481\linewidth}
        \centering
        \includegraphics[width=\linewidth]{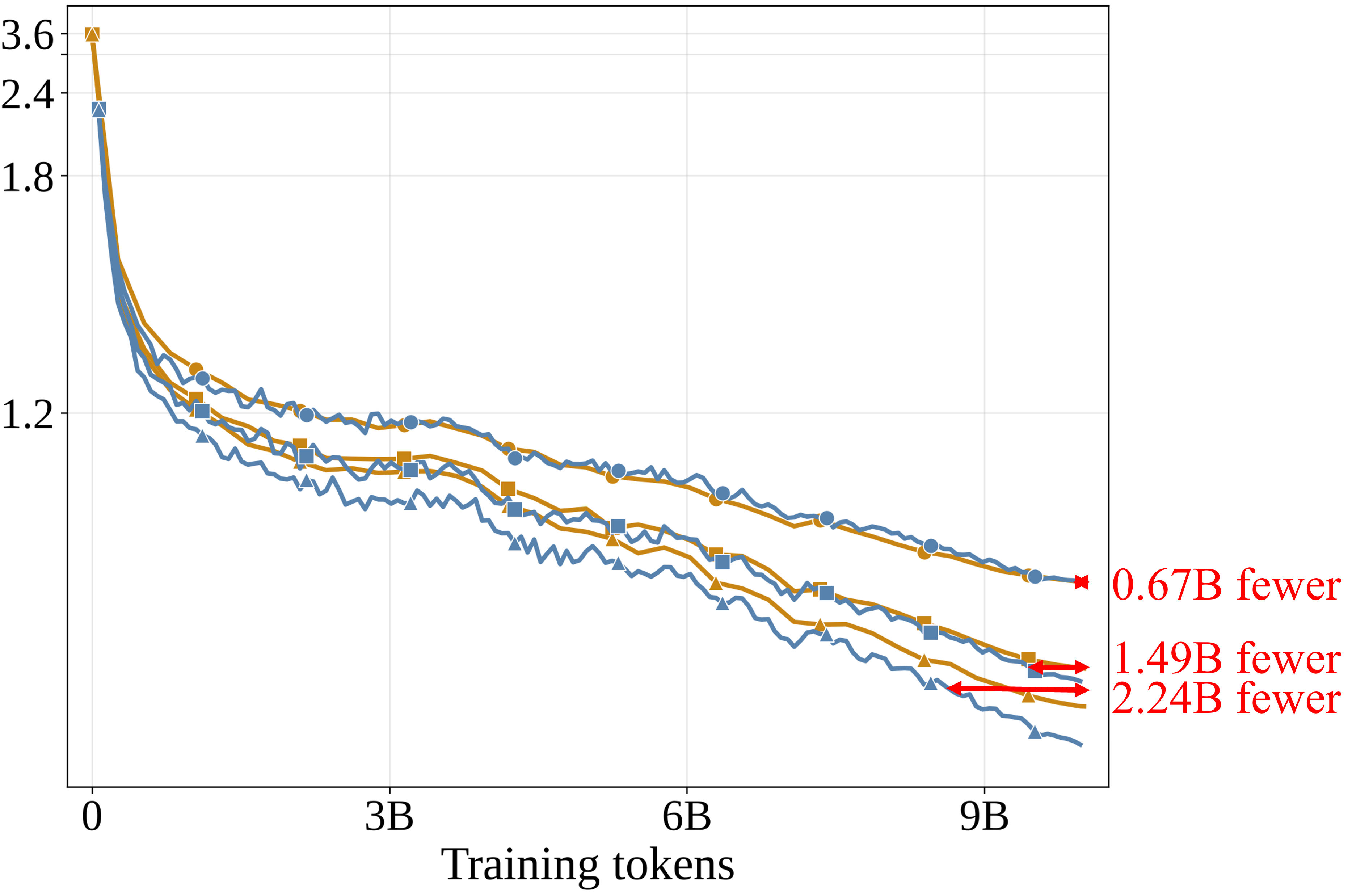}
        \caption{DCLM}
        \label{fig:pretrain_val_bpb_dclm}
    \end{subfigure}
    \begin{subfigure}[t]{0.499\linewidth}
        \centering
        \includegraphics[width=\linewidth]{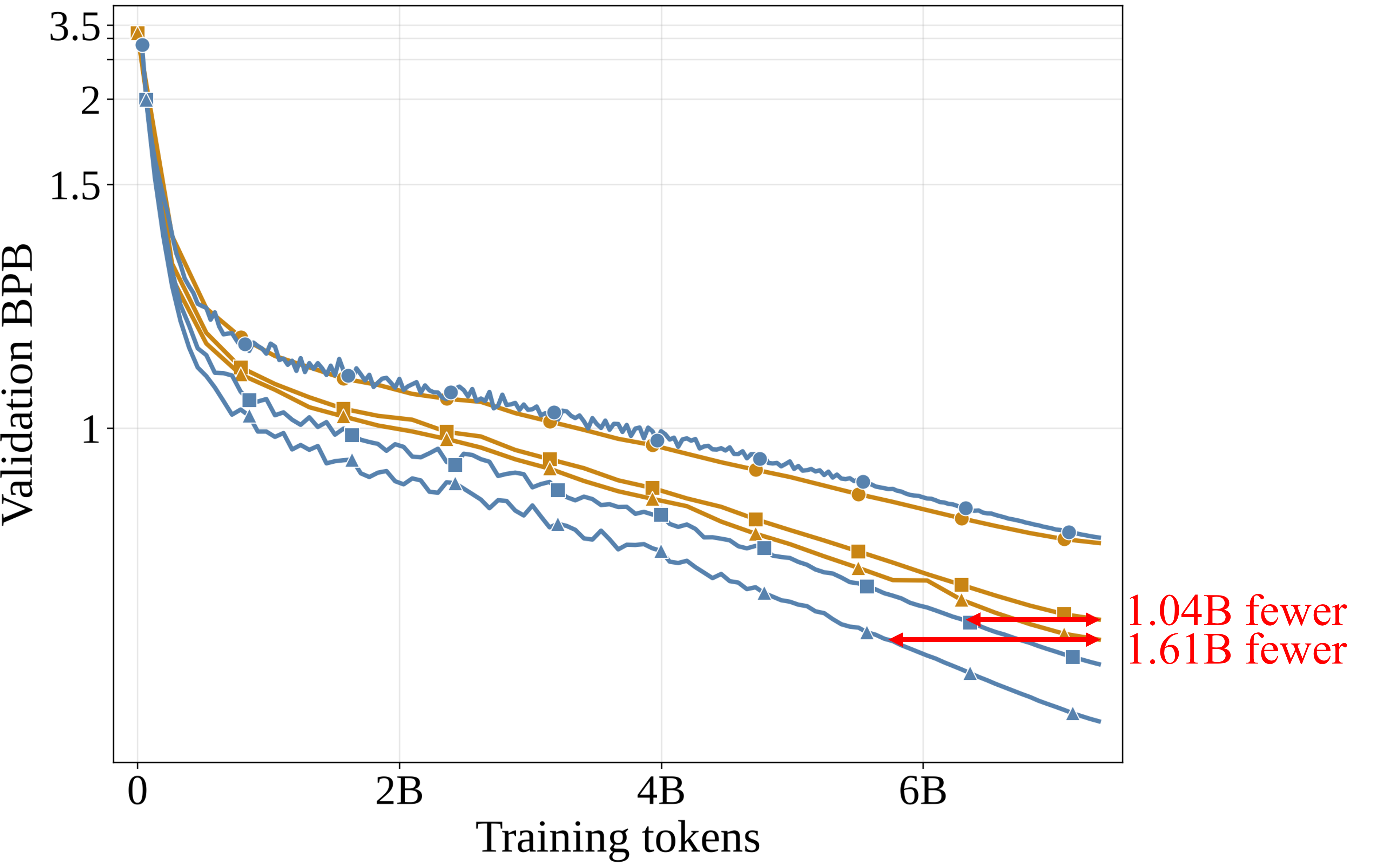}
        \caption{Fineweb}
        \label{fig:pretrain_val_bpb_fineweb}
    \end{subfigure}
    \begin{subfigure}[t]{0.481\linewidth}
        \centering
        \includegraphics[width=\linewidth]{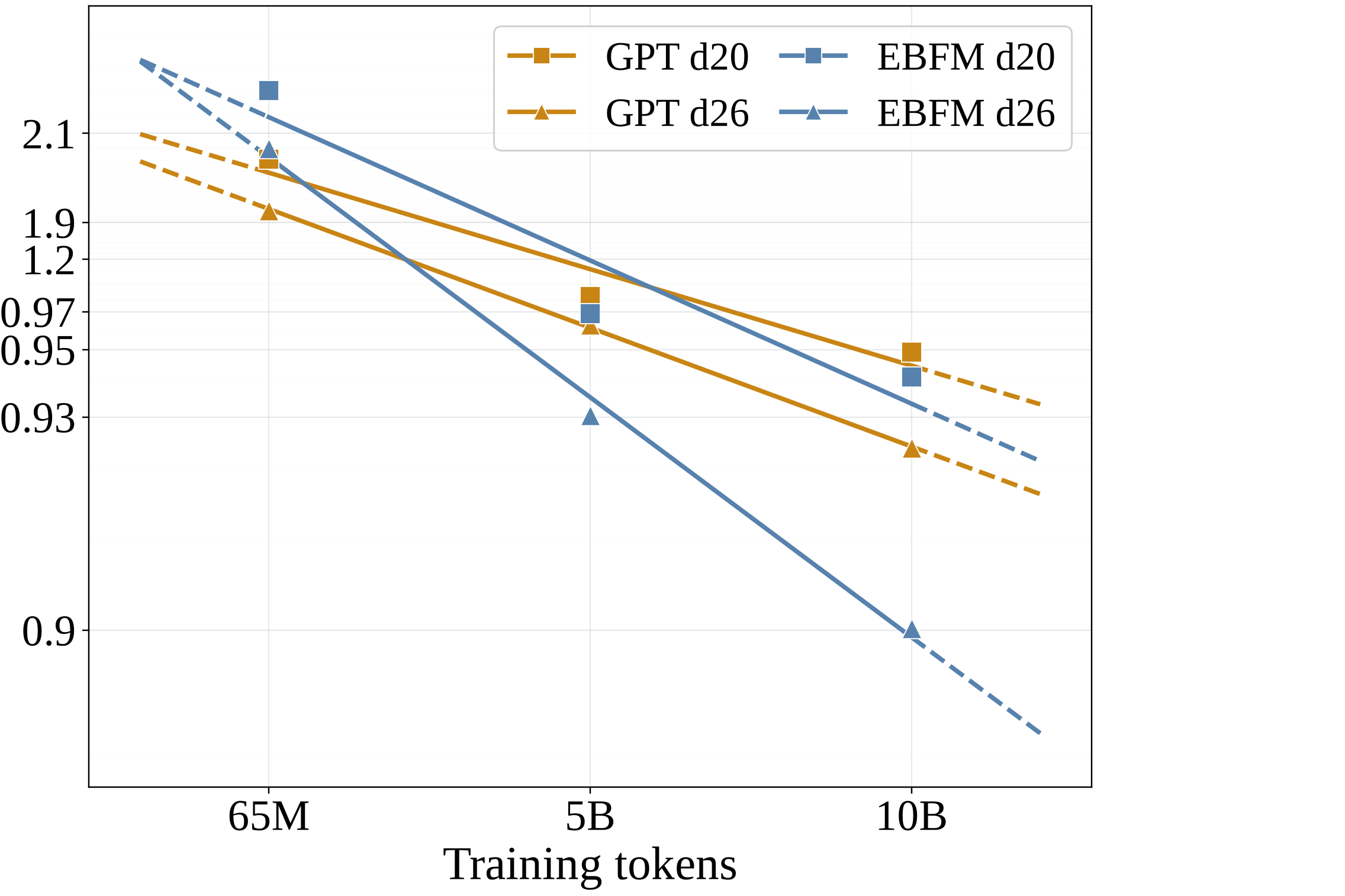}
        \caption{Scaling with training tokens}
        \label{fig:pretrain_val_bpb_scaling}
    \end{subfigure}
    \vspace{-4mm}
    \includegraphics[
        width=0.96\linewidth
    ]{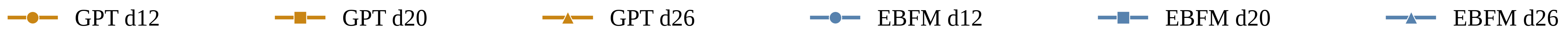}
    \vspace{2mm}

    \caption{
        Validation error during pre-training on the
        (a) Climbmix, (b) DCLM, and (c) Fineweb datasets; (d) Scaling law of \methodabbf{}s versus different training budgets
    }
    \label{fig:pretrain_val_bpb}
\end{figure}

\begin{figure}
\centering
\begin{subfigure}{0.49\linewidth}
    \centering
    \includegraphics[width=\linewidth]{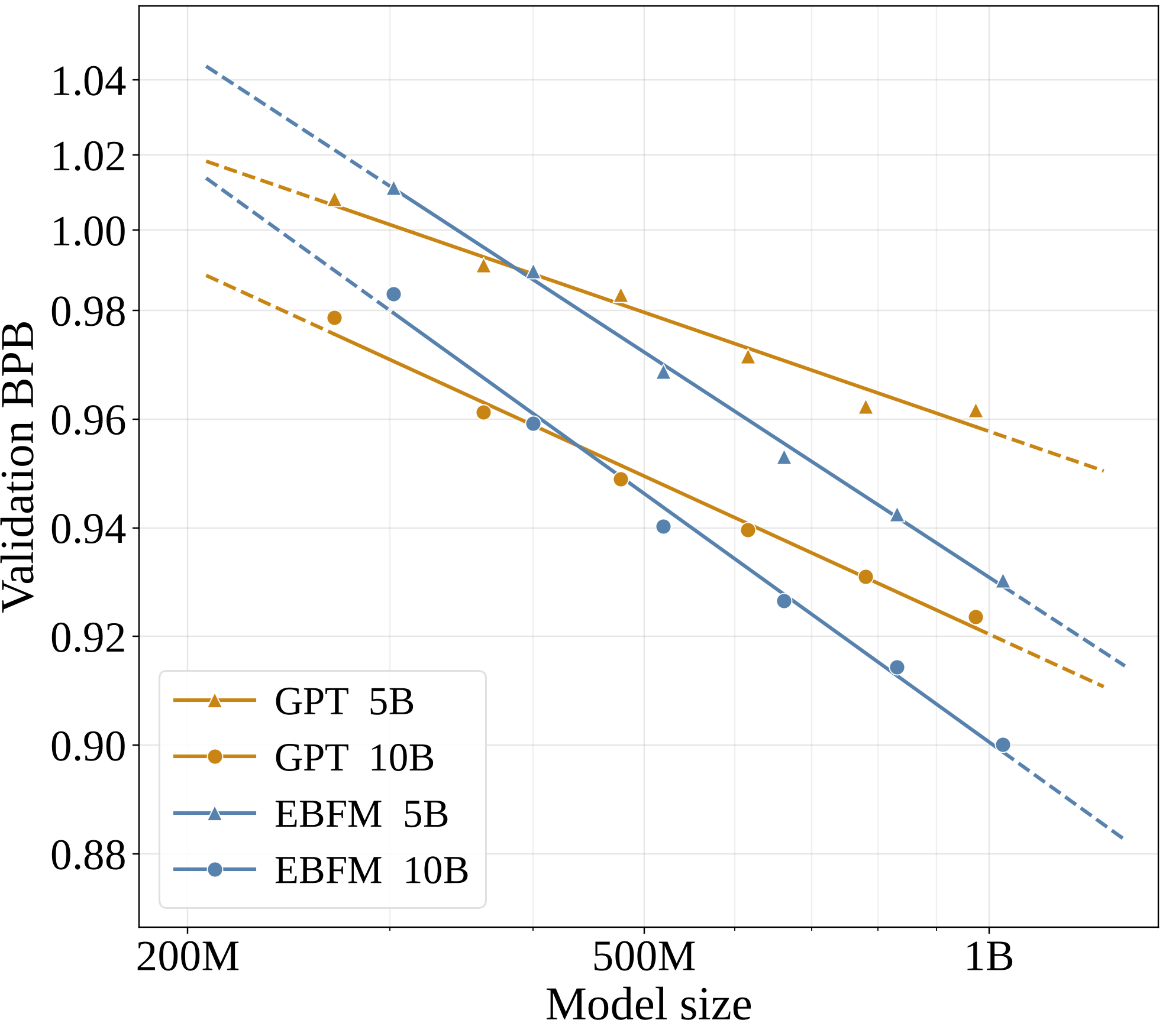}
    \caption{Minimal validation BPB
versus model depth}
\end{subfigure}
\begin{subfigure}{0.49\linewidth}
    \centering
    \includegraphics[width=\linewidth]{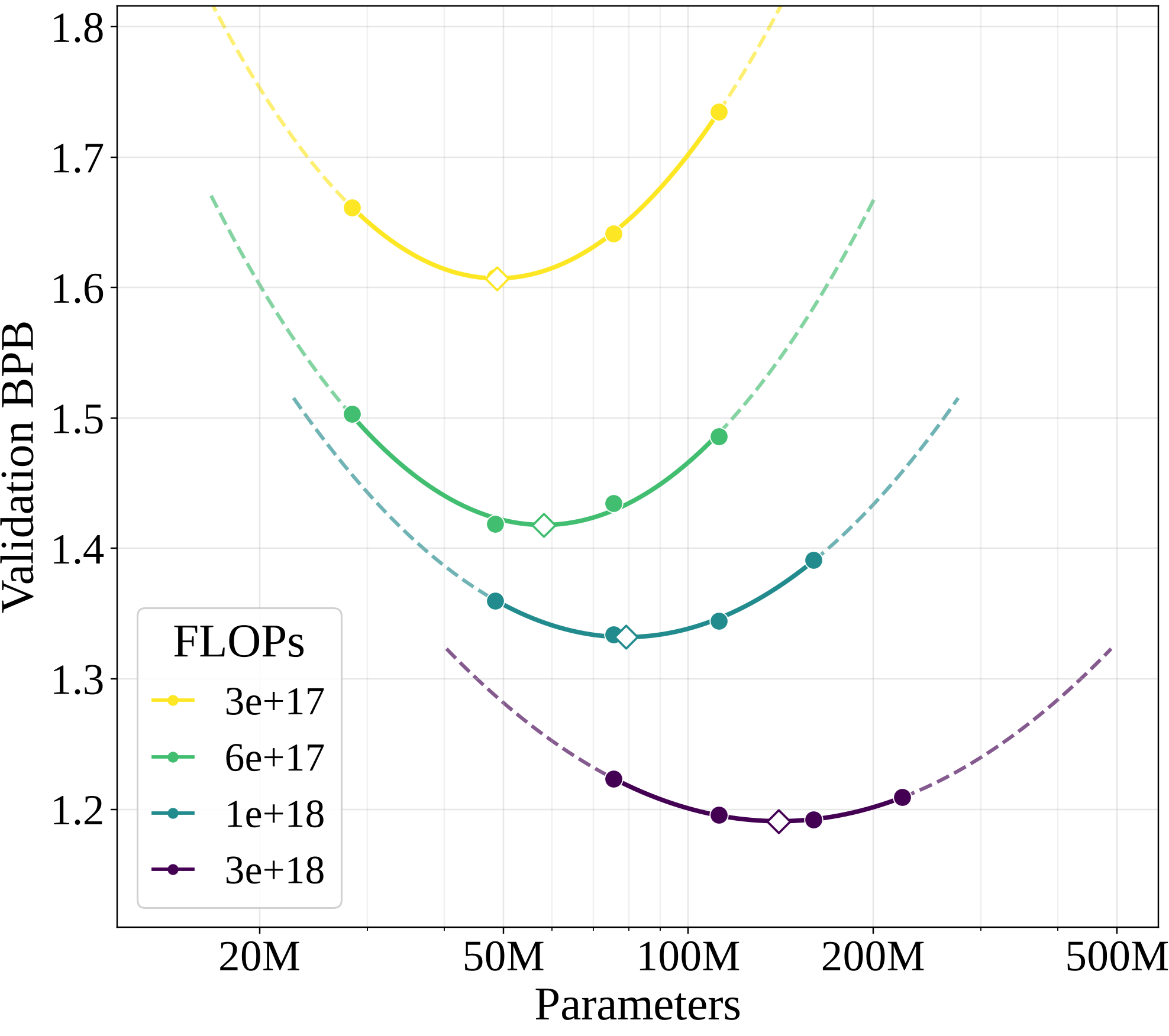}
    \caption{IsoFLOPs curves}
\end{subfigure}
\caption{\label{fig:dclm_isoflop} Scaling behaviors of \methodabbf{} with respect to model depth and training FLOPs.}
\end{figure}

\begin{figure}[t]
\centering
\begin{subfigure}{0.49\linewidth}
    \centering
    \includegraphics[width=\linewidth]{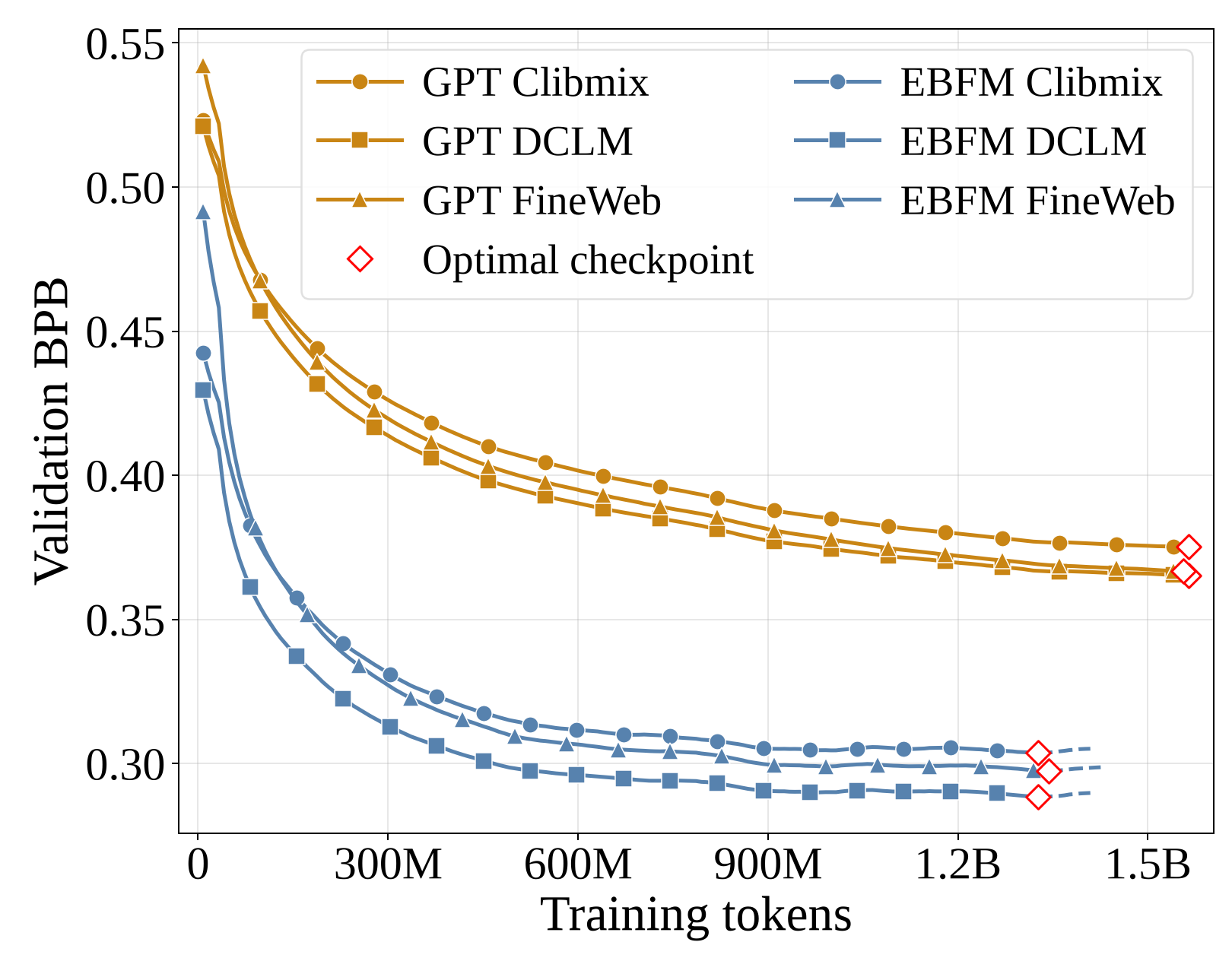}
    \caption{$d20$}
\end{subfigure}
\begin{subfigure}{0.49\linewidth}
    \centering 
    \includegraphics[width=\linewidth]{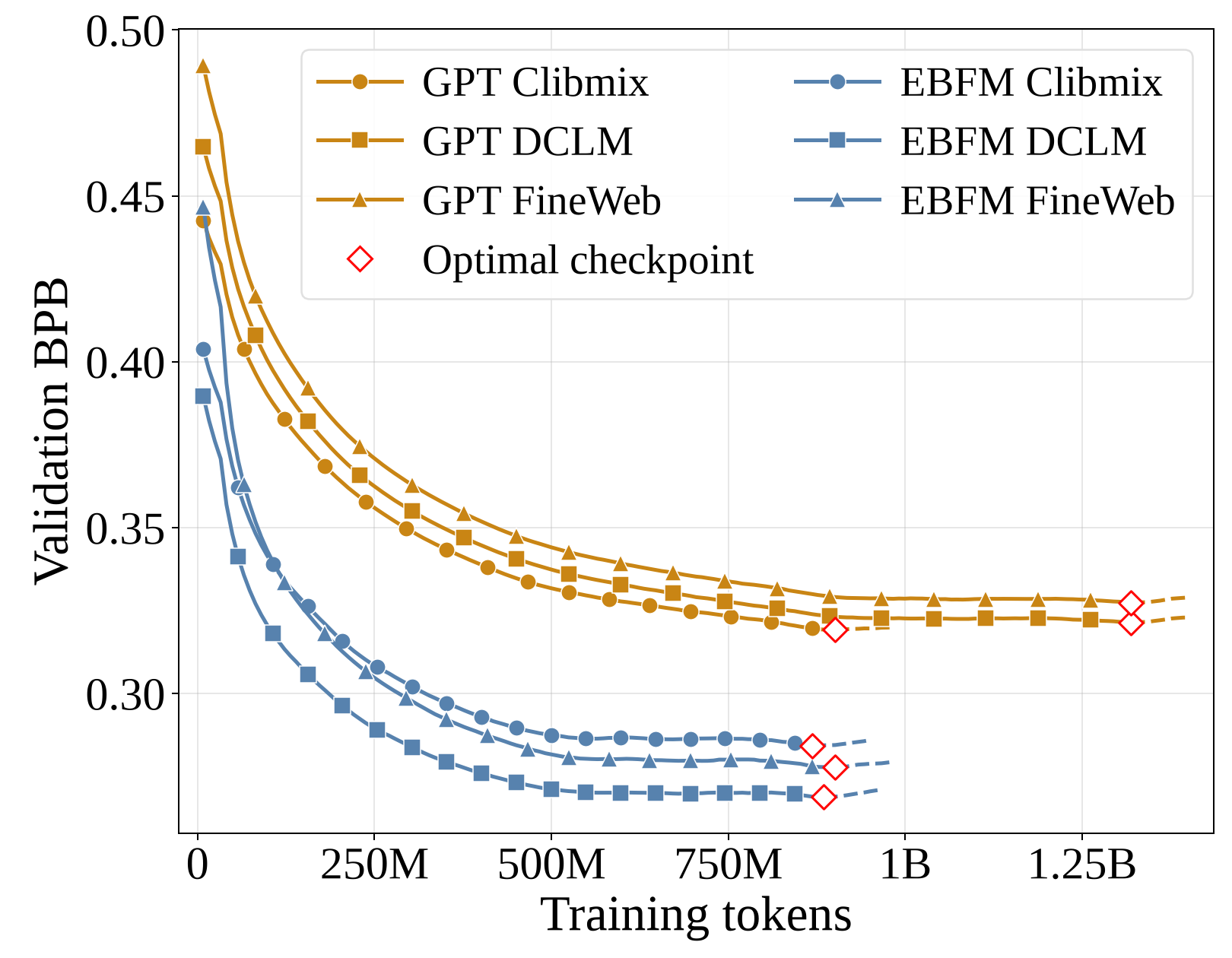}
    \caption{$d26$}
\end{subfigure}
\caption{Validation error during supervised fine-tuning. (a) $d20$ models (b) $d26$ models}
\label{fig:sft_val_bpb}
\end{figure}

\subsection{Evaluation of Scaling Law}
\label{sec:scaling}


To evaluate pre-training performance, we report the validation BPB versus training token number on the ClimbMix, DCLM, and FineWeb datasets in Figure~\ref{fig:pretrain_val_bpb}~(a), (b), and (c), respectively.
We analyze the performance in terms of training-token scaling, depth scaling, and IsoFLOP scaling as follows.

\paragraph{Data Scaling.}
Across all three corpora, \methodabbf{}s consistently outperform GPTs, with BPB decreasing more rapidly as the number of training tokens increases.
Given the same training tokens, \methodabbf{}s consistently achieves lower BPB than GPTs across different model sizes.
In details, as shown in Figure~\ref{fig:pretrain_val_bpb}~(a), on the ClimbMix dataset, \methodabbf{} with depths of $12$, $20$, and $26$ layers can match GPTs trained on 7B tokens after pre-training on $4.5$B, $3.5$B, and $5.1$B tokens, respectively.
Thus, \methodabbf{}s require $0.67$B, $1.49$B, and $2.24$B fewer training tokens, respectively.
Similarly, on the DCLM and Fineweb datasets, \methodabbf{}s significantly $1.04$B and $1.61$B fewer tokens than GPTs, respectively, to match the final performance of GPTs. 
These results demonstrate the potential of \textbf{\textit{\methodabbf{} as a data-efficient foundation architecture.}}

Moreover, Figure~\ref{fig:pretrain_val_bpb}~(d) illustrates the minimal validation error versus training tokens on the DCLM dataset with different training budgets.
As the training tokens increase from $65$M to $10$B, the validation error of \methodabbf{}s decreases faster than that of GPTs.
Thus, \textbf{\textit{\methodabbf{}s exhibit an increasing performance advantage over GPTs as the training budget grows.}}
These results establish \methodabbf{}s as a promising architecture for large-scale pre-training.

\paragraph{Parameter Scaling.}
Figure~\ref{fig:dclm_isoflop}~(a) shows the validation BPB versus model size, which are measured on the DCLM dataset with $5$B and $10$B training tokens.
As the model size increases from $268$M to $1.03$B parameters, the validation error of \methodabbf{}s reduces faster than that of GPTs.
\textbf{\textit{\methodabbf{}'s performance advantage over GPTs becomes increasingly pronounced at larger model scales.}}
This result demonstrates that \methodabbf{}s benefit more effectively from parameter scaling, highlighting their potential as a foundation architecture.



\paragraph{IsoFLOP Law.}
Figure~\ref{fig:dclm_isoflop}~(b) presents the U-shape IsoFLOP curves of \methodabbf{}s on the DCLM dataset, where each curve represents the validation BPB across different model sizes under fixed training FLOPs. 
The minimal point of each curve identifies the compute-optimal model size for \methodabbf{}s.
As the training budget increases, the optimal point shifts downward and to the right, indicating that larger \methodabbf{}s achieve lower validation error.
This trend demonstrates promising scaling behaviors of \methodabbf{}s like GPTs~\cite{hoffmann2022training}.
To be concrete, it suggests that \textbf{\textit{\methodabbf{} can consistently achieve lower validation error with increased training budget when model size and training tokens are scaled appropriately.}}
The IsoFLOP curves highlight \methodabbf{}'s potential as a foundation architecture.


\subsection{Evaluation of Supervised Fine-tuning}

We further evaluate the performance of \methodabbf{}s after supervised fine-tuning~(SFT).
The SFT takes the Nanochat-SFT dataset, which is a mixture of SmolTalk, MMLU, and GSM8K datasets~\cite{nanochat}.
For both GPTs and \methodabbf{}s, SFT is initialized from their respective optimal pre-training checkpoints, ensuring that each architecture starts from its optimal pre-trained state.
Figure~\ref{fig:sft_val_bpb} illustrates the validation error versus the training token numbers for $d20$ and $d26$ models.
During SFT, \methodabbf{}s achieve lower validation error while consuming fewer SFT tokens, indicating that their favorable pre-trained initialization enables more data-efficient fine-tuning.
This highlights the advantage of \methodabbf{}s over GPTs throughout the full training pipeline, from pre-training to SFT, establishing its potential as a foundation architecture.




\subsection{Evaluation of Question-Answering Tasks}

\begin{figure}[t]
    \begin{subfigure}{0.49\linewidth}
        \centering
        \includegraphics[width=\linewidth]{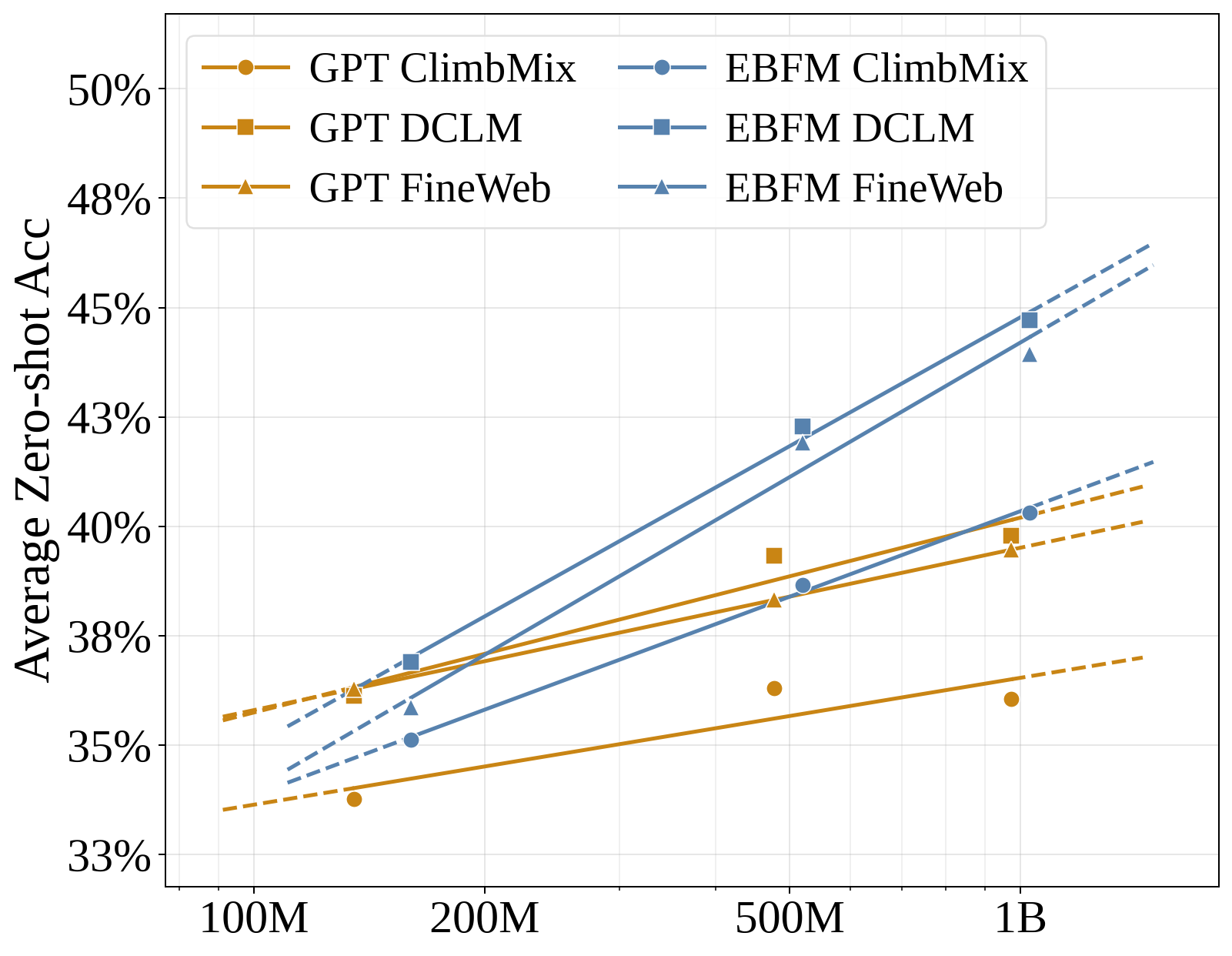}
        \caption{Model Size}
    \end{subfigure}
    \begin{subfigure}{0.49\linewidth}
        \centering
        \includegraphics[width=\linewidth]{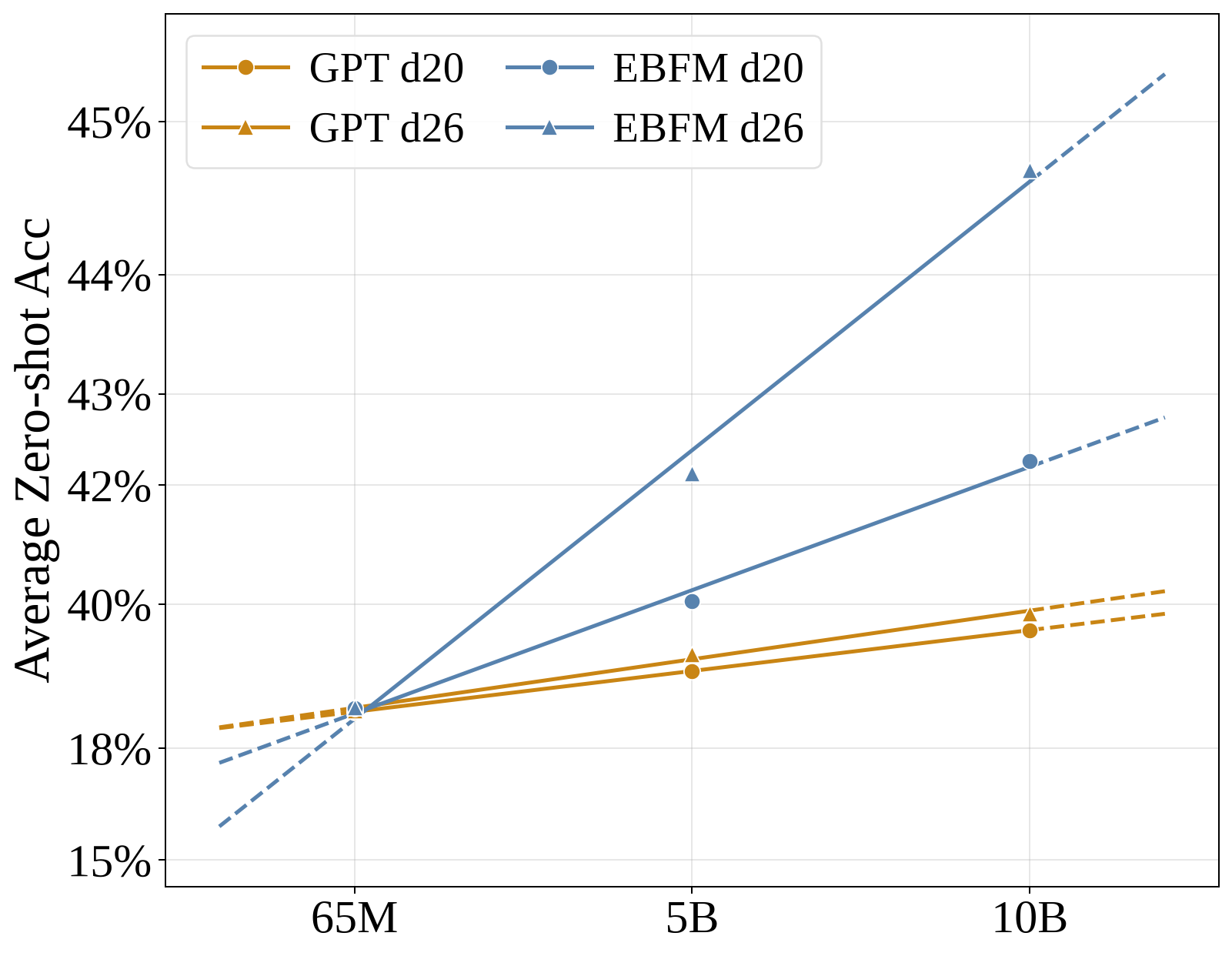}
        \caption{Training Tokens}
    \end{subfigure}
    \caption{(a) Zero-shot accuracy versus model size (b) Zero-shot accuracy versus training tokens on DCLM dataset}
    \label{fig:zero-shot-acc}
\end{figure}

We evaluate \methodabbf{}s on zero-shot question-answering~(QA).
The details of the 19 downstream QA tasks are given in Section~\ref{sec:recipe}.
According to the results in Table~\ref{tab:core_results}, \methodabbf{}s significantly outperform GPTs on most downstream tasks.
These results highlight the potential of \methodabbf{} to serve as foundation models for general-purpose tasks across diverse application scenarios.
Moreover, Figure~\ref{fig:zero-shot-acc} (a) and (b) illustrate the zero-shot accuracy versus model size and training tokens, respectively.
The zero-shot performance of \methodabbf{}s improves more rapidly with increasing model size and training tokens than that of GPTs; and the advantage of \methodabbf{}s grows at larger scales.
These results demonstrate the favorable scaling behaviors of \methodabbf{} and its stronger potential at larger scales.





\begin{table}[t]
    \centering
    \scriptsize
    \setlength{\tabcolsep}{2pt}

    \resizebox{\textwidth}{!}{%
    \begin{tabular}{lll|*{20}{c}}
    \toprule

    \multirow{2}{*}{\makecell{Pretrain \\ data}}
    & \multirow{2}{*}{Model}
    & \multicolumn{1}{c|}{\multirow{2}{*}{Depth}}
    & \multicolumn{1}{c|}{\begin{tabular}[b]{@{}c@{}}AGIEval\\[-0.6ex]\rule{1cm}{\arrayrulewidth}\end{tabular}}
    & \multicolumn{2}{c|}{\begin{tabular}[b]{@{}c@{}}ARC\\[-0.6ex]\rule{1.2cm}{\arrayrulewidth}\end{tabular}}
    & \multicolumn{5}{c|}{\begin{tabular}[b]{@{}c@{}}BIG-bench\\[-0.6ex]\rule{4cm}{\arrayrulewidth}\end{tabular}}
    & \multicolumn{1}{c|}{BoolQ}
    & \multicolumn{1}{c|}{CSQA}
    & \multicolumn{1}{c|}{COPA}
    & \multicolumn{1}{c|}{CoQA}
    & \multicolumn{2}{c|}{\begin{tabular}[b]{@{}c@{}}HellaSwag\\[-0.6ex]\rule{1.4cm}{\arrayrulewidth}\end{tabular}}
    & \multicolumn{1}{c|}{LAMBADA}
    & \multicolumn{1}{c|}{OBQA}
    & \multicolumn{1}{c|}{PIQA}
    & \multicolumn{2}{c|}{\begin{tabular}[b]{@{}c@{}}Winograd\\[-0.6ex]\rule{1.3cm}{\arrayrulewidth}\end{tabular}}
    & \multicolumn{1}{c}{Avg.}
    \\

    && \multicolumn{1}{c|}{}
    & \multicolumn{1}{c|}{LSAT-AR}
    & C
    & \multicolumn{1}{c|}{E}
    & CSAlg
    & Dyck
    & LangID
    & Ops
    & \multicolumn{1}{c|}{WikiQA}
    & \multicolumn{1}{c|}{}
    & \multicolumn{1}{c|}{}
    & \multicolumn{1}{c|}{}
    & \multicolumn{1}{c|}{}
    & 10-shot
    & \multicolumn{1}{c|}{0-shot}
    & \multicolumn{1}{c|}{}
    & \multicolumn{1}{c|}{}
    & \multicolumn{1}{c|}{}
    & WSC
    & \multicolumn{1}{c|}{WinoG}
    & \\
    \midrule
    & \multirow{3}{*}{GPT} & $d12$ & \textbf{26.96} & 23.20 & 42.80 & 40.80 & 11.20 & \textbf{25.80} & 11.90 & 28.20 & 55.60 & 21.80 & 56.00 & 15.00 & \textbf{31.00} & 30.20 & \textbf{31.20} & 26.40 & 57.00 & \textbf{56.41} & \textbf{50.20} & 33.77 \\
    &  & $d20$ & 23.91 & \textbf{26.60} & 47.60 & 41.00 & 13.20 & \textbf{26.20} & 15.71 & 34.20 & \textbf{57.60} & \textbf{28.00} & \textbf{62.00} & 19.80 & 34.00 & 34.00 & 35.80 & 25.40 & 59.60 & 59.34 & 46.00 & 36.31 \\
    &  & $d26$ & 26.52 & 27.80 & 46.60 & 37.40 & \textbf{14.20} & \textbf{26.00} & 17.14 & 37.00 & 50.40 & \textbf{25.00} & 59.00 & 20.80 & 33.40 & 32.60 & 35.00 & 27.20 & 60.00 & 57.51 & 51.60 & 36.06 \\
    \rowcolor{gray!10}
    \cellcolor{white} &  & $d12$ & 21.74 & \textbf{23.29} & \textbf{44.36} & \textbf{42.12} & \textbf{13.70} & 25.63 & \textbf{18.57} & \textbf{30.75} & \textbf{56.24} & \textbf{37.84} & \textbf{58.00} & \textbf{16.47} & 30.47 & \textbf{30.75} & 28.18 & \textbf{27.20} & \textbf{58.16} & 54.58 & 49.33 & \textbf{35.13} \\
    \rowcolor{gray!10}
    \cellcolor{white} &  & $d20$ & \textbf{24.35} & 26.02 & \textbf{50.80} & \textbf{41.59} & \textbf{17.50} & 24.83 & \textbf{21.90} & \textbf{42.55} & 56.61 & 25.72 & 61.00 & \textbf{26.92} & \textbf{36.70} & \textbf{36.60} & \textbf{38.93} & \textbf{27.00} & \textbf{61.21} & \textbf{64.47} & \textbf{50.12} & \textbf{38.67} \\
    \rowcolor{gray!10}
    \cellcolor{white}\multirow{-6}{*}{ClimbMix} & \multirow{-3}{*}{\methodabbf{}} & $d26$ & \textbf{30.00} & \textbf{29.61} & \textbf{53.87} & \textbf{40.53} & 11.90 & 25.00 & \textbf{19.52} & \textbf{47.66} & \textbf{52.97} & 23.59 & \textbf{65.00} & \textbf{26.31} & \textbf{40.16} & \textbf{40.23} & \textbf{43.12} & \textbf{30.20} & \textbf{63.76} & \textbf{66.30} & \textbf{56.27} & \textbf{40.32} \\
    & \multirow{3}{*}{GPT} & $d12$ & 24.78 & 25.00 & 47.60 & 37.40 & 12.60 & 23.80 & \textbf{13.33} & 30.20 & \textbf{55.20} & 30.40 & \textbf{63.00} & \textbf{14.00} & 32.00 & 31.60 & \textbf{38.40} & 28.40 & 64.40 & \textbf{61.54} & \textbf{52.80} & 36.13 \\
    &  & $d20$ & \textbf{26.96} & 28.00 & 53.20 & \textbf{44.40} & 12.00 & \textbf{25.80} & 13.81 & 41.80 & 55.00 & 23.00 & \textbf{66.00} & 19.20 & 38.00 & 39.00 & 44.80 & 30.80 & 68.00 & 64.84 & 52.60 & 39.33 \\
    &  & $d26$ & 23.04 & 28.40 & 54.20 & 38.60 & 18.00 & 23.60 & 17.14 & 44.00 & 52.40 & \textbf{26.00} & 61.00 & 17.60 & 42.00 & 42.20 & 48.60 & 31.00 & 70.40 & 63.74 & 54.00 & 39.79 \\
    \rowcolor{gray!10}
    \cellcolor{white} &  & $d12$ & \textbf{26.09} & \textbf{25.60} & \textbf{48.36} & \textbf{44.09} & \textbf{21.60} & \textbf{25.58} & 12.86 & \textbf{33.77} & 50.92 & \textbf{34.81} & 57.00 & 11.62 & \textbf{34.27} & \textbf{34.08} & 33.30 & \textbf{29.60} & \textbf{65.45} & 59.71 & 52.41 & \textbf{36.90} \\
    \rowcolor{gray!10}
    \cellcolor{white} &  & $d20$ & 22.61 & \textbf{29.78} & \textbf{57.74} & 41.89 & \textbf{16.30} & 24.79 & \textbf{17.62} & \textbf{48.90} & \textbf{55.02} & \textbf{32.35} & 61.00 & \textbf{22.44} & \textbf{47.92} & \textbf{47.21} & \textbf{48.09} & \textbf{33.00} & \textbf{69.91} & \textbf{73.63} & \textbf{53.28} & \textbf{42.29} \\
    \rowcolor{gray!10}
    \cellcolor{white}\multirow{-6}{*}{DCLM} & \multirow{-3}{*}{\methodabbf{}} & $d26$ & \textbf{26.96} & \textbf{32.25} & \textbf{61.78} & \textbf{44.02} & \textbf{18.30} & \textbf{24.40} & \textbf{18.10} & \textbf{54.25} & \textbf{55.29} & 25.88 & \textbf{65.00} & \textbf{26.53} & \textbf{54.70} & \textbf{53.64} & \textbf{52.94} & \textbf{35.20} & \textbf{72.42} & \textbf{71.43} & \textbf{56.35} & \textbf{44.71} \\
    & \multirow{3}{*}{GPT} & $d12$ & 21.30 & 24.60 & 53.20 & \textbf{40.80} & 9.00 & \textbf{25.60} & 11.90 & \textbf{34.40} & \textbf{56.80} & 30.80 & \textbf{62.00} & \textbf{15.20} & \textbf{35.00} & \textbf{34.80} & \textbf{31.60} & 30.80 & 62.80 & 53.48 & \textbf{55.40} & \textbf{36.29} \\
    &  & $d20$ & 22.61 & 32.60 & 57.40 & 41.20 & 10.40 & 23.20 & 13.33 & 40.00 & \textbf{50.40} & 31.00 & 62.00 & 18.80 & 40.40 & 39.40 & 33.20 & 34.00 & 66.20 & 60.07 & 51.80 & 38.32 \\
    &  & $d26$ & 24.35 & 30.60 & 60.60 & 41.00 & 8.40 & \textbf{28.80} & 15.24 & 45.40 & \textbf{58.60} & 23.80 & 63.00 & 16.80 & 40.20 & 40.80 & 36.80 & 35.00 & 67.60 & 61.90 & 51.00 & 39.47 \\
    \rowcolor{gray!10}
    \cellcolor{white} &  & $d12$ & \textbf{24.78} & \textbf{27.47} & \textbf{54.00} & 34.85 & \textbf{11.80} & 24.93 & \textbf{12.38} & 32.79 & 48.53 & \textbf{32.68} & 61.00 & 12.48 & 34.68 & 34.31 & 28.82 & \textbf{32.20} & \textbf{64.96} & \textbf{58.24} & 50.36 & 35.86 \\
    \rowcolor{gray!10}
    \cellcolor{white} &  & $d20$ & \textbf{29.13} & \textbf{35.24} & \textbf{63.22} & \textbf{44.62} & \textbf{13.20} & \textbf{24.82} & \textbf{16.67} & \textbf{48.05} & 47.37 & \textbf{35.54} & \textbf{67.00} & \textbf{21.01} & \textbf{45.39} & \textbf{45.38} & \textbf{36.46} & \textbf{35.40} & \textbf{68.06} & \textbf{66.30} & \textbf{53.59} & \textbf{41.92} \\
    \rowcolor{gray!10}
    \cellcolor{white}\multirow{-6}{*}{FineWeb} & \multirow{-3}{*}{\methodabbf{}} & $d26$ & \textbf{26.52} & \textbf{36.95} & \textbf{66.96} & \textbf{43.86} & \textbf{16.00} & 24.73 & \textbf{16.67} & \textbf{50.52} & 57.83 & \textbf{30.06} & \textbf{68.00} & \textbf{24.59} & \textbf{51.21} & \textbf{50.93} & \textbf{40.64} & \textbf{38.40} & \textbf{70.02} & \textbf{65.93} & \textbf{54.85} & \textbf{43.93} \\

\bottomrule
    \end{tabular}%
    }
    \caption{Zero-shot accuracy (\%) on 19 question-answering datasets after pre-training.}
    \label{tab:core_results}
\end{table}


\subsection{Zero-shot Generation}

\methodabbf{}s are compared with GPTs, and Diffusion Large Models~(MDLMs) for zero-shot generations.
All models are pre-trained on the OpenWebText dataset~\cite{Gokaslan2019OpenWeb}, where \methodabbf{}s and GPTs are trained for ONE epoch with $9.06$B tokens, and MDLMs are trained for TWO epochs with $18.12$B tokens.
All models are evaluated on the OpenWebText validation dataset as well as seven downstream datasets, including PTB, Wikitext, LM1B, Lambada, AG News, Pubmed, and Arxiv datasets.
The evaluation results are given in Table~\ref{tab:controllable-generation}, where \methodabbf{}s consistently outperform GPTs and MDLMs, achieving $0.02$ and $0.09$ lower BPB errors, respectively.
The evaluation results of zero-shot generation demonstrate the potential of \methodabbf{}s as a foundation architecture.

\begin{table}[t]
    \centering
    \small
    \begin{tabular}{l|c|cccccccc}
        \toprule
        Model & OpenWebText-val & PTB & Wikitext & LM1B & Lambada & AG News & Pubmed & Arxiv & Average \\
        \midrule
        \methodabbf{} & \textbf{0.93} & \textbf{1.46} & \textbf{1.10} & \textbf{1.24} & 1.53 & \textbf{1.34} & \textbf{1.34} & \textbf{1.44} & \textbf{1.35} \\
        GPT & 0.94 & 1.51 & 1.11 & 1.26 & \textbf{1.48} & 1.35 & 1.39 & 1.51 & 1.37\\
        MDLM & 1.17 & 1.53 & 1.23 & 1.38 & \textbf{1.48} & 1.52 & 1.41 & 1.50 & 1.44\\
        \bottomrule
    \end{tabular}
    \caption{\label{tab:controllable-generation} Validation BPB of \methodabbf{}s, GPTs, and MDLM pre-trained on the OpenWebText datasests, where \methodabbf{}s and GPTs are trained for only ONE epoch with $9.06$B tokens, and MDLM is trained for TWO epoches with $18.12$B tokens. Average values are estimated for zero-shot settings.}
\end{table}




\section{Open-source Release}

We fully release the training infrastructure and training logs at \href{https://github.com/datamllab/openesm}{https://github.com/datamllab/openesm}, the model weights at \href{https://huggingface.co/collections/guan-wang/openesm}{https://huggingface.co/collections/guan-wang/openesm}. 
All datasets in this work are publicly available and properly linked in our released code.
We sincerely hope the community can reproduce our results and jointly explore the potential of energy-based foundation models.

\section{Limitations under Current Ecosystem}

\paragraph{Efficiency.}


Our largest \methodabbf{} configuration is $d26$, with $1.03$B parameters. 
Pre-training efficiency remains a major bottleneck to scaling \methodabbf{} to larger sizes. 
Currently, the training of \methodabbf{} and other EBMs relies primarily on distributed data parallelism~(DDP), where each GPU must maintain a complete model backbone. 
Existing model parallel infrastructures, including FSDP, Megatron-LM, and DeepSpeed ZeRO, do not yet readily support the efficient second-order differentiation required by EBM training. 
Therefore, a single model cannot be easily sharded across multiple GPUs, which limits the model size to the memory capacity of a single H200 GPU. 
We encourage the community to extend distributed-training infrastructure to support second-order gradients and model sharding. 
Addressing this bottleneck will enable EBMs to scale to larger capacities and better performance.

\paragraph{Block-wise Generation.}
Block-wise generation enables safer and more controllable generation by jointly optimizing whole sentence under energy-based guidance.
Different from token-wise generation, which generates a locally optimal token at each step, block-wise energy evaluation enables semantic coherence and safety constraints to be assessed over a longer span.
Therefore, block-wise energy steering facilitates improving long-range contextual consistency and constraint compliance.
Extending \methodabbf{}s to block-wise generation remains an important direction for future work.

\section{Conclusion}

In this work, we propose \methodabbf{}s as a foundation architecture, and explore its scaling behaviors across general-purpose tasks.
Specifically, \methodabbf{}s perform energy steering to optimize hidden states during inference, thereby guiding token generation toward outputs with lower energy.
Moreover, we develop parallel energy steering for efficient \methodabbf{} pre-training, and further pre-train the models on three large-scale corpora.
Across multiple evaluation settings and 19 downstream tasks spanning diverse domains, \methodabbf{}s demonstrate promising scaling performance with increased model size and training cost.
These findings highlight the substantial potential of \methodabbf{}s as a foundation architecture for general-purpose tasks.
We completely release our infrastructure and model weights to enable full reproduction of our results.
We encourage the community to jointly explore the potential of energy-based foundation models.

\section{Future Directions}

We believe that the potential of \methodabbf{}s remains under-explored, and we welcome the community to investigate promising directions, including but not limited to the following:

\begin{itemize}
    \item \textbf{Supervised Fine-tuning.} Fine-tuning \methodabbf{}s on domain-specific instructions and data to acquire to strengthen their capabilities in specialized domains, such as scientific reasoning, code generation, and finance.
    Effective fine-tuning of \methodabbf{}s could further improve their adaptability to diverse applications.

    \item \textbf{Efficient Inference/Training.} \methodabbf{}s differ architecturally from GPTs, making existing efficient inference or training, such as GPTQ, Flash Attention, and Page Attention, difficult to apply. 
    Developing efficient inference and training algorithms tailored to the architecture of \methodabbf{} will be promising directions.
    
    \item \textbf{Training Infrastructure.} Developing distributed training infrastructure to scale up \methodabbf{} training. 
    In particular, future systems could support model sharding across multiple GPUs, memory-efficient computation of higher-order gradients that are comparable with frameworks such as FSDP and Megatron-LM.
    
\end{itemize}



\clearpage
\bibliographystyle{unsrt}
\bibliography{reference}

\end{document}